%% file: main.tex
\documentclass[10pt,twocolumn,letterpaper]{article}

\usepackage[pagenumbers]{cvpr} % To force page numbers, e.g. for an arXiv version

\usepackage{multirow}
\usepackage[linesnumbered,ruled,vlined]{algorithm2e}

\input{preamble}

\definecolor{cvprblue}{rgb}{0.21,0.49,0.74}

\usepackage[pagebackref,breaklinks,colorlinks,citecolor=cvprblue]{hyperref}

\usepackage{xcolor}
\newcommand{\yk}[1]{\textcolor{blue}{#1}}

\title{Hypergraph Multi-modal Large Language Model: Exploiting EEG and Eye-tracking Modalities to Evaluate Heterogeneous Responses for Video Understanding} % For ArXiv
\vspace{-15pt}
\author{
    Minghui Wu\textsuperscript{\rm 1,2,3}\thanks{Authors contributed equally to this work.} \quad
    Chenxu Zhao\textsuperscript{\rm 2,3}\footnotemark[1] \quad
    Anyang Su \textsuperscript{\rm 2,3}\footnotemark[1] \quad
    Donglin Di\textsuperscript{\rm 3} \quad
    Tianyu Fu\textsuperscript{\rm 3} \quad
    \\
    Da An \textsuperscript{\rm 3} \quad
    Min He \textsuperscript{\rm 2} \quad
    Ya Gao \textsuperscript{\rm 1,2} \quad
    Meng Ma \textsuperscript{\rm 1} \quad
    Kun Yan \textsuperscript{\rm 1}\thanks{Corresponding authors.} \quad
    Ping Wang \textsuperscript{\rm 1}\footnotemark[2]\quad
    \\
    \textsuperscript{\rm 1}Peking University \quad
    \textsuperscript{\rm 2}Mininglamp Technology \quad
    \textsuperscript{\rm 3}Shanghai Artificial Intelligence Laboratory 
    \\
    {\tt\footnotesize \{wuminghui,zhaochenxu,suanyang,hemin\}@mininglamp.com,\ gaoya@stu.pku.edu.cn, \{mameng,kyan2018,pwang\}@pku.edu.cn}
}
\begin{document}

\maketitle

\input{sec/00-abstract}    
\input{sec/01-intro}
\input{sec/02-background}
\input{sec/03-DataSet}
\input{sec/04-method}
\input{sec/05-experiment}
\input{sec/06-conclusion}
\input{sec/07-acknowledge}

\newpage
{
    \small
    \bibliographystyle{ieeenat_fullname}
    \bibliography{main}
}

\input{appendix}


% {
%     \small
%     \bibliographystyle{ieeenat_fullname}
%     \bibliography{main}
% }
\end{document}

%% file: preamble.tex
\usepackage[table]{xcolor}

\usepackage{amsmath,amsfonts,amssymb}
\usepackage{amsmath} % For advanced math typesetting
\usepackage{amsfonts} % For mathematical fonts
\usepackage{amssymb} % For mathematical symbols
\usepackage{bm}
\usepackage{algpseudocode}
\usepackage{threeparttable} 
\usepackage{subcaption}

\usepackage{todonotes}

%% file: sec/00-abstract.tex
\begin{abstract}
\vspace{-5mm}
  Understanding of video creativity and content often varies among individuals, with differences in focal points and cognitive levels across different ages, experiences, and genders. There is currently a lack of research in this area, and most existing benchmarks suffer from several drawbacks: 
  1) a limited number of modalities and 
  answers with restrictive length; 
  2) the content and scenarios within the videos are excessively monotonous, transmitting allegories and emotions that are overly simplistic. To bridge the gap to real-world applications, we introduce a large-scale \textbf{S}ubjective \textbf{R}esponse \textbf{I}ndicators for \textbf{A}dvertisement \textbf{V}ideos dataset, namely SRI-ADV. 
  Specifically, we collected real changes in Electroencephalographic (EEG) and eye-tracking regions from different demographics 
  while they viewed identical video content.
  Utilizing this multi-modal dataset, we developed tasks and protocols to analyze and evaluate the extent of cognitive understanding of video content among different users. 
  Along with the dataset, we designed a \textbf{H}ypergraph \textbf{M}ulti-modal \textbf{L}arge \textbf{L}anguage \textbf{M}odel (HMLLM) to explore the associations among different demographics, video elements, EEG and eye-tracking indicators. HMLLM could bridge semantic gaps across rich modalities and integrate information beyond different modalities to perform logical reasoning. Extensive experimental evaluations on SRI-ADV and other additional video-based generative performance benchmarks demonstrate the effectiveness of our method. The codes and dataset will be released at \url{https://github.com/mininglamp-MLLM/HMLLM}.

\end{abstract}

%% file: sec/01-intro.tex
\section{Introduction}
\label{sec:intro}

\begin{figure}
    \centering
    \includegraphics[width=\linewidth]{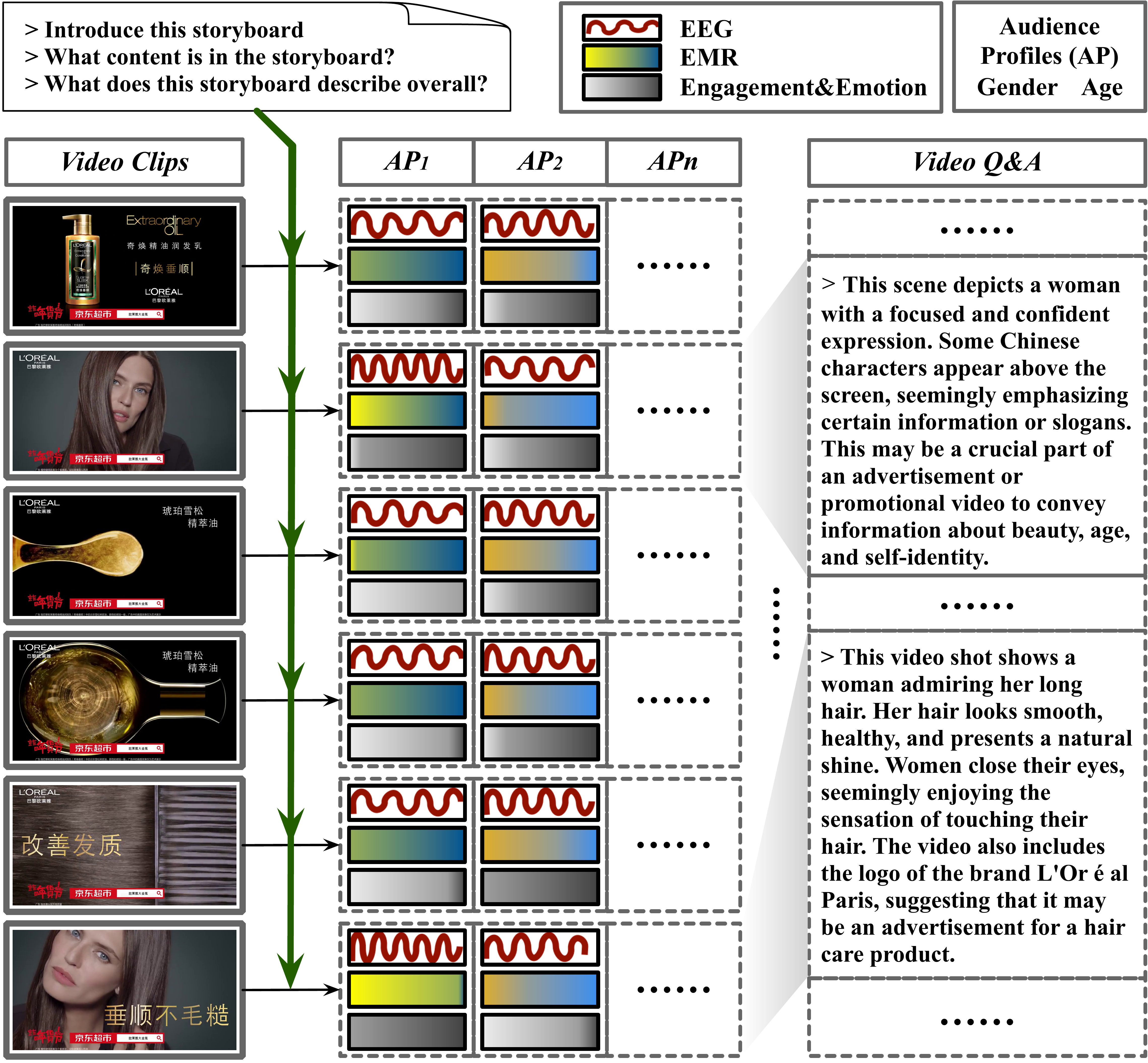}
    \caption{
    Our proposed Subjective Response Indicators for Advertisement Videos (SRI-ADV) dataset. 
    % (SRI-ADV)
    % Signals captured in real-time 
    Real-time signals captured by electroencephalographic (EEG) and eye-tracking devices reveal that Audience Profiles (AP) of varying genders and ages exhibit distinct engagements, emotions, and eye motion ratios (EMR) when exposed to 
    various scenes and elements within the same advertisement video.
    }
    \label{fig:intro}
    \vspace{-3mm}
\end{figure}

\begin{table*}[ht]
\centering
\caption{Comparison of existing VideoQ\&A datasets with ours (OE: open-ended,  MC: multiple-choice, AP: Audience Profiles).}
\label{tab:dataset_comparison}
\vspace{-3mm}
\resizebox{\textwidth}{!}{
\begin{tabular}{l|lllcrrrr}
\toprule
Datasets & Video source & Q\&A generation & Q\&A tasks & Modality & Videos & Q\&A pairs & AvgAnsLen & MedScene \\ 
\midrule
MSVD-QA \cite{xu2017VideoQuestion} & MSVD & Auto & OE & Video &1,970 & 50,505 & 1.0 & 2 \\
MSRVTT-QA \cite{xu2017VideoQuestion} & MSRVTT & Auto & OE & Video & 10,000 & \textbf{243,680} & 1.0 & 3 \\
TGIF-QA \cite{2017TGIF-QA} & TGIF & Auto\&Human & OE \& MC & Frame/Video & \textbf{56,720} & 103,919 & 1.5 & 1 \\
% MovieQA \cite{tapaswi2016movieqa} & Movies & Human & MC & Video &6,771 & 6,462 ~~ & &  \\
% Video-QA \cite{yang2003videoqa} & Jukinmedia & Auto & OE & Video &18,100 & 174,775 & ~~ &  \\
ActivityNet-QA \cite{yu2019activityqa} & ActivityNet & Human & OE & Video & 5,800 & 58,000 & 1.3 & 7 \\
Video-ChatGPT \cite{Maaz2023VideoChatGPT} & ActivityNet & Auto\&Human & OE & Video & 200 & 2,994 & 51.0 & 6 \\
% MVBench \cite{li2023mvbench}  & mutil & Automatic & MC & False & - & 428,910 & - & - \\
\midrule
SRI-ADV-QA (ours) & Custom & Auto\&Human & MC \& OE & \textbf{Video/EEG/EMR/AP} & 498 & 178,547 & \textbf{99.6} & \textbf{11} \\
\bottomrule
\end{tabular}
}
\vspace{-3mm}
\end{table*}

With the advancement of Large Language Models (LLMs) \cite{wei2022emergent} and Multi-modal Large Language Models \cite{li2020oscar,chen2020uniter,li2021supervision,jia2021scaling}, the field of video understanding has entered a new era. 
% The powerful logical reasoning capabilities of multi-modal LLMs enable not only comprehensive decoding of the explicit factors within videos but also allow for the inference of implicit content behind these explicit factors through the knowledge and experience learned by LLMs. 
The advanced logical reasoning abilities of multi-modal LLMs facilitate a thorough analysis of explicit elements within videos. Moreover, these models can deduce the underlying implicit content of these explicit factors, leveraging the knowledge and experience acquired by LLMs.
Existing benchmarks for video content question-and-answering, such as \cite{xu2017VideoQuestion,xu2017VideoQuestion,2017TGIF-QA,yu2019activityqa,Maaz2023VideoChatGPT}, provide a rich set of instruction labels. Alternatively, they exhibit several deficiencies as illustrated in Table \ref{tab:dataset_comparison}: 1) the video content itself is overly simplistic, often only involving objective, explicit factors, which does not support the exploration of deeper levels of video creativity and implicit factors. We utilize the \textbf{MedScene} metric to evaluate this issue, where \textbf{MedScene} denotes the median number of scene across all videos in the dataset. A higher number of scenes indicates greater complexity in video content; 2) the number of modalities included in these datasets are limited, generally confined to videos and frames; 3) the instruction labels concerning the length of answers are restricted to certain predetermined options, failing to assess the divergent and analytical abilities of LLMs. We utilize the \textbf{AvgAnsLen} to evaluate this issue, where \textbf{AvgAnsLen} represents the average text length of the answer portion across all Q\&A pairs in the dataset. 
To address the issues mentioned above, we have prepared an extensive collection of content-rich advertisement videos, accompanied by a more comprehensive set of modality labels.

In the burgeoning field of cognitive neuroscience, the exploration of how individuals perceive and interpret video content has opened new avenues for understanding the intricate interplay between brain activity and media interaction \cite{vajda1950mathematical}. Recent advancements in multi-modal data analysis have underscored the importance of leveraging diverse physiological signals to gain insights into the cognitive and emotional states of viewers \cite{khandelwal2023large}. Among these, Electroencephalographic (EEG) signals with their high temporal resolution, provide a direct measure of brain activity \cite{roy2019deep}, capturing the nuanced and dynamic changes in cognitive states as individuals engage with video content. These signals embody the electrical manifestations of the brain's complex neural dynamics, offering insights into the emotional and cognitive processes underpinning video content interpretation \cite{rolls2018brain}. 
%By analyzing EEG patterns, researchers can identify specific neural markers associated with different cognitive and emotional responses, enabling a more nuanced understanding of how video content is processed and perceived by various individuals.

Inspired by the aforementioned context, we have utilized EEG and eye-tracking apparatus to collect and record the EEG and eye movement responses of individuals across various ages, genders, and professions while watching the same advertisement video. We aggregated this information into modality labels, introducing a novel, large-scale benchmark: the Subjective Response Indicators for Advertisement Videos dataset, namely \textbf{SRI-ADV}. As illustrated in Figure \ref{fig:intro}, our proposed dataset captures the subjective reactions of individuals watching videos through EEG and eye-tracking devices, fills the gaps in the video understanding domain regarding the assessment of video appeal and implicit factors. 
How to effectively leveraging these multi-modal labels to uncover the latent associations among the modalities becomes the cornerstone for addressing deeper challenges in video understanding. 

Graph-based methodologies exhibit superiority in exploring the associations among features, particularly hypergraphs, extending beyond traditional graph theory, offer a powerful framework for representing complex relationships in data \cite{berge1984hypergraphs}. In the context of video content analysis, hypergraphs can encapsulate the intricate associations among video elements, EEG signals, and eye-tracking data, allowing for the modeling of higher-order interactions that are not capturable through simple pairwise connections. 
%This capability enables the exploration of the multi-dimensional nature of cognitive responses to video content, reflecting the complexity of human cognition and its susceptibility to a wide array of stimuli. The Hypergraph approach facilitates the identification of subtle patterns and correlations within the data, shedding light on the underlying mechanisms driving diverse reactions to video content.

Utilizing the multi-modal information of the SRI-ADV dataset, coupled with the superiority of constructing associative features through hypergraph, we proposed a Hypergraph Multi-modal Large Language Model (\textbf{HMLLM}), integrating information from disparate modalities to perform logical reasoning and semantic analysis. By leveraging the rich information encoded in video content, along with EEG and eye-tracking data, HMLLM can bridge semantic gaps across modalities, offering a comprehensive understanding of the cognitive processes involved in video content interpretation. 
%This integration enables the model to synthesize insights from multiple data sources, enhancing its ability to predict and analyze the heterogeneous reactions of different user groups to the same video content.

The main contributions can be summarized as follows:

1. Introduction of a novel large-scale benchmark dataset: the Subjective Response Indicators for Advertisement Videos (SRI-ADV) dataset, a large-scale benchmark that captures real-time EEG and eye-tracking data from a diverse demographic while they watch advertisement videos. This dataset fills a significant gap in the field of video understanding by providing rich modality information and a comprehensive set of question-and-answer (Q\&A) pairs that allow for the assessment of video creativity and implicit factors. 

2. Development of the Hypergraph Multi-modal Large Language Model (HMLLM): we have developed a novel HMLLM that leverages the complex relationships among video elements, EEG signals, and eye-tracking data encapsulated in hypergraphs. 
%This model represents a significant advancement in the field of video content analysis, enabling the exploration of higher-order interactions that are beyond the reach of simple pairwise connections. By integrating information from disparate modalities, HMLLM bridges semantic gaps across rich modalities, enhancing its capability for logical reasoning and semantic analysis. This integration allows for a comprehensive understanding of the cognitive processes involved in video content interpretation.

3. Extensive experimental evaluations demonstrating our method's effectiveness: through rigorous experimental evaluations conducted on the SRI-ADV dataset and additional video Q\&A datasets, we have demonstrated the effectiveness of our HMLLM. 

%% file: sec/02-background.tex
\section{Background}
\label{sec:background}

\subsection{Video Understanding}
Video understanding aims to create algorithms that allow machines to interpret videos with the same expertise as humans. Meanwhile, video emotion recognition~\cite{li2023towards}\cite{pan2022representation}\cite{zhang2023weakly} emphasizes the interplay between the emotions conveyed by the video and the viewer responses, collectively forming a critical component of video understanding.
Most existing works focus on modeling objective and tangible visual properties of videos~\cite{diba2020large}, particularly in action recognition~\cite{Two-stream,I3D,TSN,slowfast,3dfusion,lstm1,i3d+lstm,timesformer,vtn,vivit} and temporal action localization/detection~\cite{dap,tag,single-tad1}. 
However, the need for content recommendation systems has spurred research into subjective and intangible aspects (e.g. the appeal and memorability of content~\cite{constantin2019computational}), where various semantically rich information are considered~\cite{cohendet2019videomem,azcona2020predicting,zhao2021multi,newman2020multimodal}.
% This includes analysis of the emotions conveyed by the video content, which plays a critical role in understanding viewer engagement and perception.

Compared with the above work, we present a new large-scale dataset filled with content-rich advertisement videos. This dataset includes a wider range of labels that cover both tangible and intangible aspects of content. Leveraging this dataset, we introduce an advanced hypergraph multi-modal large language model. This model is designed to simultaneously process various modalities, enabling it to conduct logical reasoning and perform in-depth semantic analysis of video content.

\subsection{EEG-Based Emotion Recognition}
% EEG signals offer several benefits, such as high temporal resolution, the ability to detect subtle emotional nuances, and providing spatial information about brain regions involved in emotion processing
% ~\cite{eeg}
% \cite{buzsaki2004neuronal}. These advantages enhance the understanding of neural mechanisms behind emotional disorders and aid in developing targeted interventions. To address the complexity of EEG signals related to emotions, researchers have extensively explored various machine learning and deep learning techniques.
Electroencephalography (EEG) signals provide detailed insights into brain activity related to emotions, offering spatial information on specific brain regions involved~\cite{buzsaki2004neuronal}. 
The Arousal-Valence model~\cite{10.1037/h0077714} is a key framework for classifying emotions along two dimensions. Xiaolin et al~\cite{9326323} explored various features to enhance the emotion recognition model. However, there's a shift towards deep learning due to the limitations of machine learning. The dynamical graph convolutional neural network (DGCNN)~\cite{8320798} was proposed to learn discriminative EEG features and interrelationships among EEG channels. Some works have moved towards multi-modal learning for robust results in EEG signal recognition tasks, such as integrating physiological signals in the multi-modal framework to enhance emotion recognition accuracy~\cite{wu2022investigating} and employing proper windowing and channel selection to avoid relying on the full length of EEG and EOG signals for classification~\cite{cai2023emotion}.
Furthermore, advancements in neuromorphic computing led to the use of Spiking Neural Networks (SNN)~\cite{SNN} for classifying spatiotemporal EEG data with lower computational requirements~\cite{kasabov2015spiking}.

\subsection{Multi-modal Large Language Models}
Multi-modal Large Language Models (MLLMs), primarily serving as vision-language models, transform images or videos into texts. These models are mainly divided into two categories: traditional large-scale pretraining~\cite{git,blip2,fromage} and instruction tuning using pre-trained LLMs~\cite{minigpt4,llaVa,mplug-owl}. 
The first category comprises models that blend a visual encoder with a language model, either developed from scratch or based on pre-existing models, possibly including a trainable module to bridge the two modalities. Utilizing auto-regressive loss for text generation, these models are training on extensive image-text datasets, including image-text pairs~\cite{git,blip2,fromage,kosmos-1} and image-text sequence instances~\cite{flamingo}. 
The second category, drawing inspiration from instruction-tuning techniques used in MLLMs~\cite{instructgpt,achiam2023gpt4}, incorporates instruction-following data to enhance MLLMs' zero- and few-shot learning abilities~\cite{minigpt4,llaVa,mplug-owl,instructblip}. A notable example is LLaVA~\cite{llaVa}, which employs a simple projection matrix to link a pre-trained visual encoder with an LLM, focusing initially on pre-training for feature alignment before comprehensive end-to-end fine-tuning. Some other works extend to video understanding by connecting video encoders to MLLMs~\cite{videochat,videogpt,videollama,Video-llava}. 
In addition to models that focus on combining images or videos with text, there are projects that incorporate even more types of data, like speech, audio, and sensor information~\cite{speechgpt,pandagpt,nextgpt,imagebind}. 

\subsection{Hypergraph Learning}
A hypergraph includes vertices and hyperedges, where hyperedges can connect multiple vertices. This structure is more adaptable and effective for representing complex relationships in data than traditional graphs~\cite{0-gao2020hypergraph}.
Methods for creating hypergraphs fall into two groups: explicit and implicit. Explicit methods directly use the data structure to form hyperedges, like connecting vertices with shared attributes~\cite{1-huang2015learning,2-joslyn2019high}. 
Implicit methods, however, infer hyperedges from data without clear high-order links, utilizing approaches based on distance~\cite{3-gao20123} or representations
% ~\cite{4-wang2015visual,5-liu2016elastic,6-jin2019robust}.
~\cite{4-wang2015visual,5-liu2016elastic}.
Unlike static structures, some methods allow for hypergraph structure optimization, adjusting it during the learning phase. This involves adaptively changing weights on hyperedges~\cite{7-gao2012visual} or sub-hypergraphs~\cite{9-zhang2018inductive} to improve learning outcomes.
Recent advancements have introduced deep hypergraph representation learning, a new approach that mainly divides into spectral~\cite{HNN,HyperGCN} and spatial~\cite{HyperSAGE,UniGNN} categories based on how hypergraph convolution operator is defined.

%% file: sec/03-DataSet.tex
\section{SRI-RAV Dataset}
\label{sec:DataSet}

\begin{figure*}
    \centering
    \includegraphics[width=1.0\linewidth]{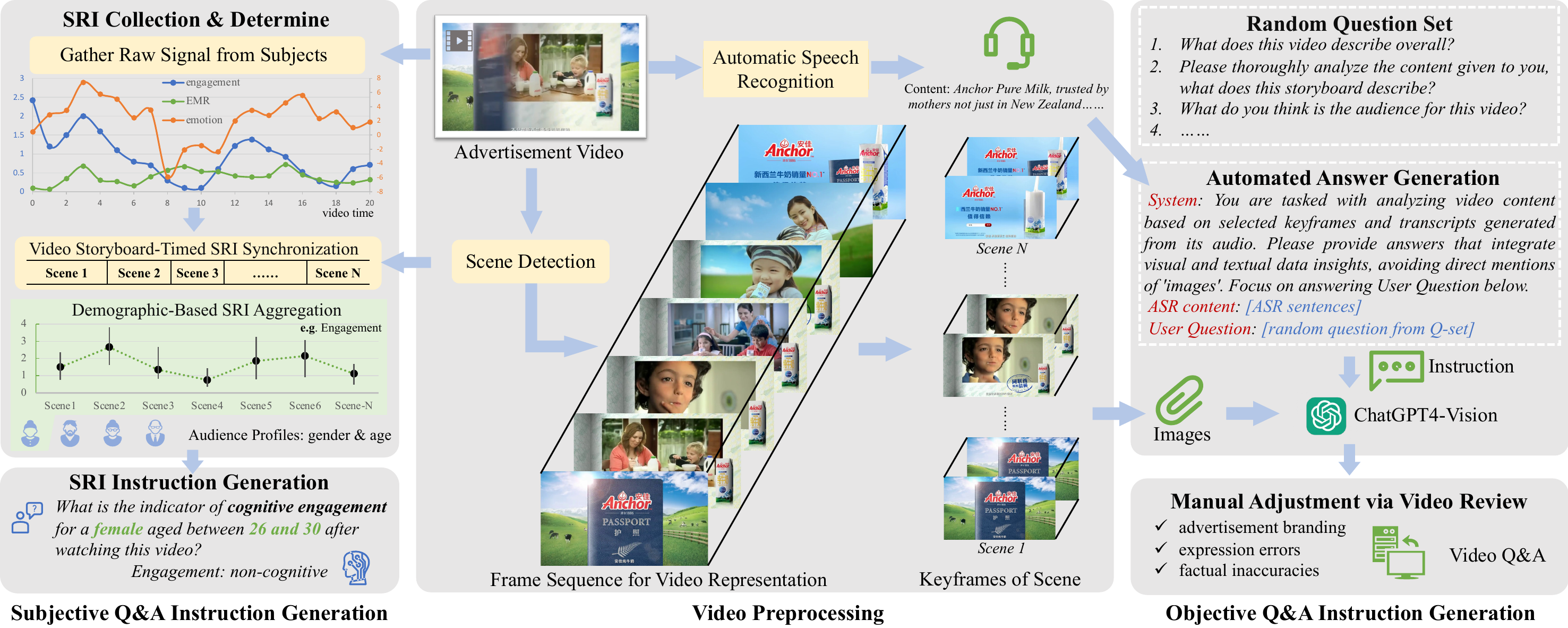}
    \caption{
    Generation pipeline of SRI-ADV dataset. The left side of this figure illustrates the process of SRI data collection, computation, and amalgamation. This involves acquiring raw signals from subjects, processing signals by video scenes, and pooling data from subjects with similar demographic profiles to obtain aggregated subjective response indicators and instruction for language models. The middle section depicts the video preprocessing with Frame Sequence for Video Representation (FSVR) by scene detection and Automatic Speech Recognition (ASR) for videos. On the right side, we present our proposed semi-automated video Q\&A generation process, which leverages both video storyboarding from FSVR and dialogue text from ASR. This integration enriches video content comprehension, thereby facilitating both Subjectivity and Objectivity Tasks.
    }
    \label{fig:QA-generation}
\end{figure*}

In this section, we present the Subjective Response Indicators for Advertisement Videos~(\textbf{SRI-ADV}) dataset. The SRI-ADV dataset not only focuses on the Objectivity Task typically found in traditional video Q\&A datasets but also meticulously collects subjective indices to enhance the richness. It encompasses a wide array of advertisement videos across different industries. To capture a diverse set of responses, we enlisted participants from various cities throughout Mainland China. These participants are equipped with EEG devices, enabling us to monitor their brainwave activities and eye motion ratios (EMR) in real-time while watching the advertisements. 
% The gathered data was then analyzed to create a benchmark for classifying brainwave and EMR responses, which is elaborated upon in Sections \ref{sec:FSVR} and \ref{sec:SRI}.
The collected data is subsequently analyzed to establish a benchmark for the classification of brainwave and EMR responses, which is elaborated in Sections \ref{sec:FSVR} and \ref{sec:SRI}.

Additionally, the SRI-ADV dataset includes an extensive video Q\&A section to provide objective insights into the ads, facilitating model training and subjective index assessment. The task definition and protocol of our dataset are outlined in Section \ref{sec:AVQA} and \ref{sec:protocol}.
% Beyond the subjective response index portion of the SRI-ADV dataset, we have also expanded it to include a comprehensive video question-and-answer (Q\&A) component. This segment provides detailed, objective descriptions of the advertisement videos, which not only facilitates a deeper understanding of the video content but also enhances the training and optimization of models for assessing subjective indices. Details of the associated video Q\&A challenge are elaborated in Section \ref{sec:AVQA}. Utilizing the distinctive attributes of our dataset, we established a benchmark, as detailed in Section \ref{sec:protocol}, for evaluating and comparing model performance.

% ============
\subsection{Frame Sequence for Video Representation}
\label{sec:FSVR}
% 论述数据集特点，为什么要做分镜、数据特点、分镜算法
The SRI-ADV dataset features Chinese advertising videos from diverse fields such as food and beverages, household items, consumer electronics, cultural tourism, software, and automobiles. It comprises 498 curated landscape videos sourced from online platforms and TV commercial ads, each running for 15-30 seconds. 

In this study, we introduce the Frame Sequence for Video Representation (\textbf{FSVR}) strategy to preprocess advertisement videos, as depicted in the middle part of Figure \ref{fig:QA-generation}. 
We enhance the video scene sensitivity by integrating the AdaptiveDetector \footnote{\url{https://www.scenedetect.com/}} for FSVR with specific parameters: \texttt{adaptive\_threshold = 2}, \texttt{min\_scene\_len = 10}, \texttt{window\_width = 2}.
%We have integrated the AdaptiveDetector of scene detection \footnote{\url{https://www.scenedetect.com/}} with the following parameters to enhance the video scene sensitivity: \texttt{adaptive\_threshold = 2}, \texttt{min\_scene\_len = 10}, \texttt{window\_width = 2}. %The AdaptiveDetector algorithm employs a dual-pass approach, where it initially uses the ContentDetector to assess frame content variances against a predefined threshold, followed by the application of a rolling average to the outcomes, assisting in the reduction of false detections particularly during camera movements. 
In the case of advertisement videos with frequent scene changes, the scene detection algorithm captures more information compared to average frame capture methods. Moreover, it is invaluable in minimizing redundant frames in videos primarily composed of static scenes.

By employing FSVR, we are able to deconstruct the temporal sequence of advertisement video frames, achieving capabilities including modality signal alignment, video content understanding, and semi--automated Q\&A instruction generation. 
%As demonstrated in the Table \ref{tab:dataset_comparison}, we introduce the median scene value as \textbf{MedScene}, comparing it with other video datasets to illustrate that our video data encompasses a broader dimension and range of scenes.

% ============
\subsection{Subjectivity: SRI Collection \& Classification}
\label{sec:SRI}
% 需要查找中文文献、脑电计算公式、项目原始资料进行补充
We developed a sophisticated system for collecting subjective indicators. Each participant watches a series of advertisement videos using the device described in the appendix. During this process, we synchronously gather EEG and eye-tracking data, along with anonymized demographic details.
% Participants watch a series of advertisement videos on this device, during which we synchronously gather EEG and eye-tracking data, along with anonymized demographic details. 
% Our study encompasses an extensive participant base of more than 4,600 individuals, offering a broad demographic representation. This includes a diverse group from white-collar workers and civil servants to students and freelancers, covering various age and income levels.
Our study includes over 4,600 participants, ensuring a wide demographic representation. The diverse participant base spans white-collar workers, civil servants, students, and freelancers across various age groups and income brackets.

The raw EEG signals are characterized by parameters such as \(\alpha_1, \alpha_2\ldots\beta_2, \beta_3\) \cite{kaur2015eeg,klimesch1999eeg}, which is detailed in the appendix. 
Given the unique demands of advertisement video analysis, we pinpointed two pivotal EEG metrics: engagement and emotion, 
as delineated by Equation 1 and Equation 2, respectively.
\begin{align}
    EN_t &= \left(\beta_2 + \beta_3\right) / \left(\alpha_3 + \alpha_2 + \beta_2 + \beta_3\right), \\
    EM_t &= \left( \alpha_3 - \alpha_2\right) / \left(\alpha_3 + \alpha_2\right) \times 100,
\end{align}
where \(EN_t\) and \(EM_t\) represent the engagement and emotion of the individual user at the sampling moment, respectively. Furthermore, we tracked eye movement data, defining the Eye Movement Ratio (\(EMR_t\)) as the proportion of time the participant's gaze fixates on the display relative to the total video duration.

% \begin{equation}
% \label{eq:engagement_once}
% EN_t = \left(\beta_2 + \beta_3\right) / \left(\alpha_3 + \alpha_2 + \beta_2 + \beta_3\right)\yk{,}  %\frac{}
% \end{equation}

% \begin{equation}
% \label{eq:emotion_once}
% EM_t = \left( \alpha_3 - \alpha_2\right) / \left(\alpha_3 + \alpha_2\right) \times 100\yk{,}
% \end{equation}

% The SRI Collection \& Determine workflow is illustrated on the left side of Figure \ref{fig:QA-generation}. It involves collecting raw signal from subjects at a rate of 10-30 recordings per second, demonstrating high-frequency data capture on a sub-second scale.Recognizing that video content evolves through scenes rather than seconds, we implemented Video Storyboard-Timed SRI Synchronization. This approach yields time-averaged, participant-specific Subjective Response Indicators (SRI). Subsequently, we aggregated SRIs from individuals sharing demographic traits into cohesive groups. These groups comprise 5-20 participants of the same gender, with an age variance of no more than 5 years. Examples of such groups include \{female, under 20 years\}, \{male, 26-30 years\}, and \{female, 46-50 years\}. This grouping method, termed Demographic-Based SRI Aggregation, described in Equation \ref{eq:average}.
The SRI Collection \& Determine workflow, depicted on the left of Figure \ref{fig:QA-generation}, captures sub-second high-frequency raw signals data. To align with video content's scene-based evolution, Video Storyboard-Timed SRI Synchronization was adopted, producing time-averaged and participant-specific SRIs. 
% These SRIs were then grouped by demographic characteristics 
Demographic characteristics then grouped these SRIs into units of 5-20 same-gender participants with a maximum age difference of 5 years, such as \{female, <20\}, \{male, 26-30\}, and \{female, 46-50\}, as Demographic-Based SRI Aggregation in Equation \ref{eq:average}.
\begin{equation} 
\label{eq:average}
\vspace{-1mm}
\bar{X} = \frac{1}{P \cdot N} \sum_{i=1}^{P} \sum_{j=1}^{N} X_{p_i,t_j}, \forall p_i \in [AP], \forall t_j \in [t_1, t_2],
\end{equation}
where \(X_{p_i,t_j}\) denotes the original Subjective Response Indicators such as  \(EN_t\), \(EM_t\), and \(EMR_t\). Each indicator associated with discrete values for participant \(p_i\) at specific timestamps \(t_j\), where \(t_j\) signifies the effective sampling moment instances within the video storyboard timeframe from FSVR in Section \ref{sec:FSVR}.

% We have successfully derived numerical indicators for collective subjective evaluations from user groups. To advance our analysis of these indicators qualitatively, we conducted a comprehensive examination of the data distribution at the Audience Profile level, grouping it into various segments. As illustrated in Table \ref{tab:SRI-ADV-dis}, the final row visually summarises the distribution for each SRI under consideration. We divide Engagement into two distinct classes, utilizing both the Leuven Engagement Scale (LES) and the observed data distribution as our guide. On the other hand, the indicators for Emotion and EMR exhibit a normal distribution pattern and are therefore segmented into three nearly equal-sized categories. Finally, by employing a mix of random question sampling and a structured approach, we devised the SRI Instruction Generation protocol detailed in Table \ref{tab:Protocol}.

For quantitative analysis, we meticulously examined data distribution across various Audience Profile segments. Engagement was categorized into two groups using the Leuven Engagement Scale (LES) and its distribution. Emotion and EMR indicators, which followed normal distributions, were divided into three equal categories. For detailed data distribution, refer to the appendix. The SRI Instruction Generation protocol is detailed in Table \ref{tab:Protocol}.

% ============
\subsection{Objectivity: Semi-automated Generation}
\label{sec:AVQA}
% 半监督标注流程（画图，选择两段视频进行对比）,公开，对比其他数据集，
In addition to subjective indicators from Audience Profiles, we developed a semi-automated annotation pipeline for ChatGPT4-Vision (GPT4V) to obtain Objective Video Q\&A, depicted in Figure \ref{fig:QA-generation}. Although GPT4V cannot process videos, it supports multiple consecutive key-frames simultaneously. Based on FSVR in Video Preprocessing, we extracted middle frames from each shot as key-frames that effectively represent the entire video. During each invocation of GPT4V to automatically generate answers, questions are selected randomly from the Random Question Set to enhance the diversity of Q\&A sessions, along with providing ASR text and FSVR key-frames. Lastly, annotators were carefully selected to manually refine objective Q\&A instruction from Automated Answer Generation, addressing issues like advertisement branding, expression errors, and factual inaccuracies.

% During the Video Preprocessing phase, we performed scene detection as FSVR and Automatic Speech Recognition (ASR) processing. Subsequently, we developed semi-supervised labeling instructions pipeline for the ChatGPT4-Vision. Although the GPT4-V does not support video files, it can upload multiple images simultaneously. Based on the results of FSVR, we extracted middle frames from each shot that effectively represent the entire video. During each invocation, we randomly select questions from the Random Question Set to enhance the diversity of Q\&A sessions, providing both ASR and key-frames. Lastly, we meticulously selected fifteen annotators trained to manually modify the semi-supervised generated dialogue outcomes. Modifications include categories such as advertisement branding, expression errors, and factual inaccuracies.

% Through the process described above, we have enabled Q\&A annotation and inference processes in multi-modal large language model that do not support videos. This approach is not only applicable in the semi-supervised annotation but can also be employed during the inference stage, especially for ImageLLM. 

% ============
\subsection{Data Overview, Tasks and Protocols }
\label{sec:protocol}

Based on the processing presented in Sections \ref{sec:SRI} and \ref{sec:AVQA}, SRI-ADV is categorized into subjectivity and objectivity tasks. The subjectivity task examines the SRI, whereas the objectivity task is dedicated to the qualitative analysis of video content and audience perception. %As shown in Figure \ref{tab:dataset_comparison}, our videos feature extended response lengths and more scenes, considering multiple modalities such as Video, EEG, EMR and Audience Profiles, thereby providing a comprehensive dataset for analysis.
As shown in Table \ref{tab:Protocol}, we present the tasks, protocols, and instructions associated with the SRI-ADV dataset. 

Task 1, entitled \textbf{Subjectivity}, is formulated as a classification task, aimed at examining the influence of video content and user characteristics on the SRI. We develope two experimental protocols to guide this investigation. The first protocol (\textbf{P1}) is designed to assess the SRI ability of a broad audience, involving the analysis of average responses across different videos. This approach is relatively straightforward. 
% . This entails analyzing the average response of users across various videos and scenes, which is relatively straightforward. 
The second protocol (\textbf{P2}) introduces a layer of complexity by focusing on the SRI discernment of particular user demographics. This necessitates a comprehensive examination of how response patterns fluctuate among diverse user cohorts.

Task 2, designated as \textbf{Objectivity}, mirrors the video Q\&A tasks prevalent in prior datasets, as described in Section \ref{sec:AVQA}. Building on the method outlined in \cite{Maaz2023VideoChatGPT}, this study conducts a supervised analysis of the answers generated, assessing their accuracy and allocating scores. This approach is designed to objectively ascertain the narrative coherence of the advertisement content and its efficacy in captivating the target audiences.

\begin{table}[t]
\centering
\caption{Task and Protocol of SRI-ADV Dataset. In Task1, Protocol1 
 (\textbf{P1}) targets a broad audience. Protocol2 (\textbf{P2}), based on P1, contains SRI to Audience Profiles.}
\label{tab:Protocol}
\begin{tabular}{l|cc|c}
\toprule
\textbf{Task Name} & \multicolumn{2}{c|}{\textbf{1. Subjectivity}} & \textbf{2. Objectivity} \\
\midrule
\textbf{Eva. Form} & \multicolumn{2}{c|}{Multi-classification} & Text generation \\
\textbf{Train Video} & \multicolumn{2}{c|}{426} & 426 \\
\textbf{Test Video} & \multicolumn{2}{c|}{72} & 72 \\
\textbf{Train Q\&A} & \multicolumn{2}{c|}{145,107} & 5762 \\
\textbf{Test Protocol} & P1 & P2 & -- \\
\textbf{Test Q\&A} & 2,640 & 26,724 & 954 \\
\bottomrule
\end{tabular}
\vspace{-1mm}
\end{table}

%% file: sec/04-method.tex
\section{Method}
\label{sec:method}
%1. Overview：首先，写明白为啥我们选择用超图多模态大模型作为整体框架，从我们提出的数据集特点出发，数据集包含视频和视频caption标签，观看者的脑电眼动标签，观看者的属性标签，数据集本身即是多模态数据，本质上我们要跨越多模态间的语义鸿沟来构建这些标签之间的联系，所以我们使用超图建立关联特征辅助多模态大模型跨过多种模态进行推理。其次，写明白创新点就在于框架，涉及到在复杂模态之间建立关联性的场景都可以使用我们的框架，描述清楚方法图; 2. 超图部分：写明超图特点，写清楚公式； 3. MLLM部分：写清楚多模态部分结构，公式，训练过程等；
%%方法部分可以主要参考文献：abode的一篇论文LARGE CONTENT AND BEHAVIOR MODELS TO UNDERSTAND, SIMULATE, AND OPTIMIZE CONTENT AND BEHAVIOR
%%实验部分：1. 消融实验；2. proposed benchmark上不同任务和协议的实验；3. 某一公开数据集实验（可选）

\begin{figure*}
    \centering
    \includegraphics[width=1.0\linewidth]{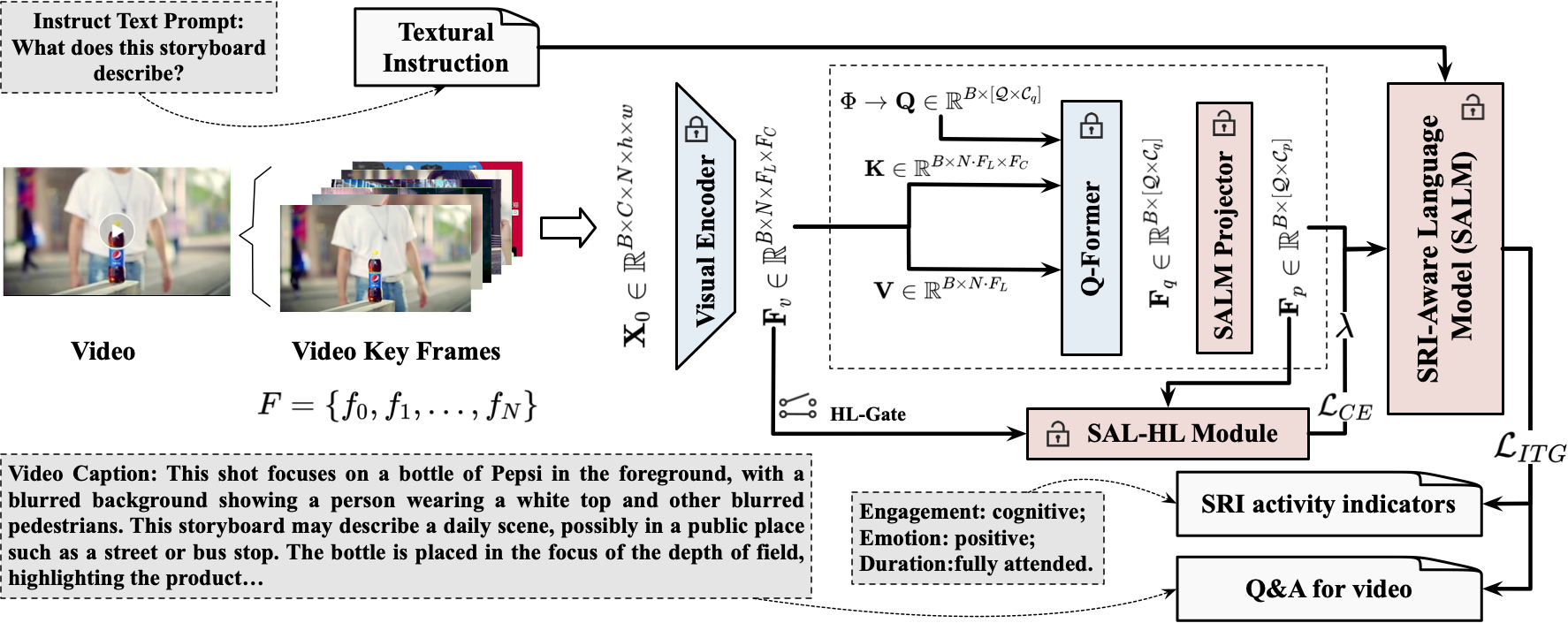}
    \caption{
    Overview of the Hypergraph Multi-modal Large Language Model (HMLLM).
    %a sophisticated system integrating advanced language and vision models with SRI data to enhance video understanding. 
    The architecture comprises a suite of pre-trained models, including a ``Visual Encoder'', ``Q-Former'', and the ``SRI-Aware Language Model (SALM)'', which are initially frozen and subsequently fine-tuned through strategic training procedures. More importantly, our model incorporates a designed ``SRI-Aware Language Hypergraph Learning (SAL-HL)'' module that is trained de novo via a combined loss function. During inference, the HMLLM generates SRI and Q\&A responses tailored to the video content, thereby providing a deeper level of engagement and comprehension.
    }
    \label{fig:method_framework}
    \vspace{-2mm}
\end{figure*}

This section elaborates on the Hypergraph Multi-modal Large Language Model (HMLLM), an approach designed to intelligently process video clips and textual prompts for generating contextually relevant text, including Subjective Response Indicators (SRI).
Central to our methodology are several key components as depicted in Figure~\ref{fig:method_framework}: Visual Encoder, Query Former (Q-Former), SALM Projector, SRI-Aware Language Model (SALM), and SAL-HL Module.
% all 
All components mentioned above synergistically orchestrated across two primary phases: SALM Warm-Up and SAL-HL Fine-Tuning, as depicted in our model architecture (refer to Figure \ref{fig:method_framework}).
The pseudocode in the appendix illustrates the detailed training process.

\subsection{SALM Warm Up}
We begin by detailing the initial stage.
The approach ingests brief video clips and corresponding textual prompts, extracting key frames from the videos using a predefined, static extraction strategy, which can be either random or uniformly distributed. 
These key frames are represented as $F = \{f_0, f_1, \ldots, f_N\}$, with $N$ signifying the 
% total
number of extracted frames.
% The video keyframes 
These key frames are then pre-processed to form the initial data matrix, denoted by $\mathbf{X}_0 \in \mathbb{R}^{B\times C\times N \times h \times w}$, where $B$, $C$, $N$, $h$, and $w$ correspond to the batch size, color channels (RGB), the number of keyframes, and the resized dimensions of the frames, respectively.
The initial data matrix $\mathbf{X}_0$ is fed into a pre-trained visual encoder to yield initial visual representations, expressed as $\mathbf{F}_v \in \mathbb{R}^{B\times N \times F_L \times F_C}$, with $F_L$ and $F_C$ representing the length and channels of features, respectively.

During the first training phase, the ``Hypergraph Learning Gate (HL-Gate)'' remains inactive while the Q-Former and SALM are warmed up.
The visual features $\mathbf{F}_v$ are then input into the frozen Q-Former as the Key 
% ($\mathbf{K} \in \mathbb{R}^{B\times N\cdot F_L \times F_C}$) 
($\mathbf{K} \in \mathbb{R}^{B\times (N \times F_L) \times F_C}$)
and Value
% ($\mathbf{V} \in \mathbb{R}^{B\times N\cdot F_L}$) 
($\mathbf{V} \in \mathbb{R}^{B\times (N \times F_L)}$) 
for the attention mechanism.
The Query in the Q-Former is initialized as either a random or null set, represented by 
% $\mathbf{Q} \in \mathbb{R}^{B \times \left [\mathcal{Q} \times \mathcal{C}_q \right ]}$,
$\mathbf{Q} \in \mathbb{R}^{B \times (\mathcal{Q} \times \mathcal{C}_q)}$,
where $\mathcal{Q} \times \mathcal{C}_q$ are the predefined hyper-parameters for the length and channels of the query.
Subsequently, we introduce an ``SALM Projector'', a multi-layer perceptron that follows the Q-Former, capable of reshaping the data and introducing additional learning parameters into the model.
The output of projector is denoted as 
% $\mathbf{F}_p \in \mathbb{R}^{B\times \left[ \mathcal{Q} \times \mathcal{C}_p \right]}$, 
$\mathbf{F}_p \in \mathbb{R}^{B\times (\mathcal{Q} \times \mathcal{C}_p )}$, 
with $\mathcal{C}_p$ being another predefined hyper-parameter.
The SRI-Aware Language Model (SALM) is then engaged, taking the output of the SALM Projector ($\mathbf{F}_p$) and the corresponding textual instructions as inputs during the initial warm-up training stage.
The SALM is trained using the Image-grounded Text Generation (ITG) loss function \cite{blip2} ($\mathcal{L}_{ITG}$), which instructs the Q-Former to generate text conditioned on the input images.
The goal of the ITG loss is to minimize the difference between the generated caption $\widetilde{\mathcal{Y}}_{qa} \gets \text{SALM}(\mathbf{F}_p, T)$ and the ground-truth caption $\mathbf{Y}_{gt}$.
This is typically achieved using a cross-entropy loss computed over the words or tokens in the caption.
The ITG loss function can be mathematically represented as:
% \begin{equation}
%     \mathcal{L}_{ITG} = - \Sigma \left(\log \mathbb{P}({\mathbf{Y}_{gt}}_i | {\mathbf{Y}_{gt}}_1, \cdots, {\mathbf{Y}_{gt}}_{i-1}, \mathbf{F}_v) \right)
% \end{equation}
\begin{equation}
    \mathcal{L}_{ITG} = - \sum \left(\log \mathbb{P}({\mathbf{Y}_{gt}}_i | {\mathbf{Y}_{gt}}_1, \cdots, {\mathbf{Y}_{gt}}_{i-1}, \mathbf{F}_v) \right)
\end{equation}
where $\mathbb{P}({\mathbf{Y}_{gt}}_i | {\mathbf{Y}_{gt}}_1, \cdots, {\mathbf{Y}_{gt}}_{i-1}, \mathbf{F}_v)$ denotes the probability of generating the $i$-th word in the caption given the previous words and the visual features extracted from the image.
% The sum is over all the words or tokens 
The summation encompasses all words or tokens
in the ground-truth caption. 

In our approach, we integrate specific strategies from BLIP2 \cite{blip2} to address the limitation of Q-Former architecture on direct interactions between the image encoder and text tokens.
%By utilizing queries to extract and channel essential information from images to text via self-attention layers, we ensure comprehensive visual information capture for text generation.
%To fine-tune the dynamics between queries and text, we adopt a multi-modal causal self-attention mask, allowing queries to interact among themselves but restricting their direct contact with text tokens. Conversely, text tokens are designed to access all queries and their preceding tokens.
Following the aforementioned training procedure, the SALM Projector and SALM are adequately warmed up, preparing them for subsequent fine-tuning optimization.

\subsection{SAL-HL Fine-tune}

In the subsequent fine-tuning phase, the hypergraph learning gate (HL-Gate) is activated, and the hypergraph learning module (SAL-HL) undergoes training in tandem with the fine-tuning of the SRI-Aware Language Model (SALM).
As delineated in Figure \ref{fig:method_framework}, the SAL-HL module receives the initial visual features ($\mathbf{F}_v$) and the representations of the projected frames ($\mathbf{F}_p$) produced by the warmed SALM Projector as inputs.

The SAL-HL module initiates the process by merging these two feature sets (i.e., $\mathbf{F}_p, \mathbf{F}_v$) and then pooling them to generate frame-level representations ($\mathbf{F}_{frame\_level}$).
%, formulated as:
This process is formulated as:
\begin{equation}
    \mathbf{F}_{frame\_level} = \text{Pool}\left( \text{Feature\_Mixer} \left ( \mathbf{F}_p  \mathbf{F}_v \right) \right).
\end{equation}

The $\textit{Feature\_Mixer}$ denotes the mixing operation between two feature matrices, which can be implemented as a multi-layer perceptron (MLP).
Each frame, denoted as $f_i$ for $i \in \left[ 0, N\right]$, is considered a vertex ($\mathcal{V}$) within the hypergraph structure ($\mathcal{G}$), which facilitates the establishment of high-order relationships among the frames.
The construction of the hypergraph entails the application of a clustering algorithm that links frames with similar latent visual features.
After constructing the hypergraph, we proceed to train the Hypergraph Neural Network (HGNN) \cite{HNN} in parallel with the Structured Attention Layer Mechanism (SALM).
This process is mathematically formulated as follows:
\begin{equation}
    \widetilde{\mathcal{Y}}_{sri}=\sigma\left(\mathbf{D}_v^{-1 / 2} \mathbf{H} \mathbf{W} \mathbf{D}_e^{-1} \mathbf{H}^{\top} \mathbf{D}_v^{-1 / 2} \cdot \mathbf{F}_{frame\_level} \cdot \boldsymbol{\mathbf{\Theta}}\right),
\end{equation}
% In this equation, 
where $\widetilde{\mathcal{Y}}_{sri}$ represents the predicted output from the SALM-enhanced HGNN, and
% The function 
$\sigma$ denotes a non-linear activation function, which introduces the necessary non-linearity into the model for capturing complex patterns.
$\mathbf{D}_e \in \mathbb{R}^{E \times E}$, $\mathbf{D}_v \in \mathbb{R}^{N \times N}$, and $\mathbf{W} \in \mathbb{R}^{E \times E}$ denote the diagonal degree matrix of hyperedges, the degree matrix of vertices, and weight matrix of hyperedges, respectively.
$\mathbf{H} \in \mathbb{R}^{N \times E}$ signifies the incidence matrix that connects hyperedges to their constituent vertices.
% The term $\mathbf{W}$ denotes the weight matrix associated with the HGNN, which is learned during training to optimize performance.
$\sigma(\cdot)$ denotes the nonlinear activation function (e.g., $\mathrm{LeakyReLU}(\cdot)$).
$\mathbf{\Theta}$ is a diagonal matrix representing the learnable parameters updated by the \textit{Cross\_Entropy} loss function in the fine-tuning loop.
It functions similarly to a multilayer perceptron (MLP) layer.
Finally, $\mathbf{F}_{frame\_level}$ represents the input feature vectors associated with the vertices of the hypergraph.
By employing this formulation, we effectively leverage the structural complexity of the hypergraph to enhance the learning capabilities of the HGNN, enabling it to capture and utilize the intricate relationships inherent within the data.
This joint training regimen integrates two loss functions: the Cross-Entropy loss ($\mathcal{L}_{CE}$) and the Image-grounded Text Generation (ITG) loss from the prior stage.
The combined loss function is expressed as:
\begin{equation}
    \label{eq:loss}
    \mathcal{L} = \mathcal{L}_{ITG} + \lambda \cdot \mathcal{L}_{CE},
\end{equation}
where $\lambda$ is a hyperparameter that balances the influence of the Cross-Entropy loss and the ITG loss on the overall optimization process. 
This composite loss function ensures that the model not only generates text that is grounded in the visual content but also adheres to the learned high-order relationships within the hypergraph structure.
% thereby enhancing the ability of the model to capture complex interactions and dependencies among the video frames.
This enhances the model's capability to capture intricate interactions and dependencies among video frames.

%% file: sec/05-experiment.tex
\section{Experiment}
%5.1 Dataset&Protocols：介绍实验中所有用到的数据集和协议，其中我们自己的数据集和协议不用过多介绍，可以直接说参考Dataset章节；
% Performance Metrics: 介绍实验中用到的衡量性能的指标，引用文献简单介绍即可；
% 5.2 Optimization：我们自己方法的实验参数；
%以上三个写到一起，可以叫overview；

%Intra Testing：分为两部分，
%第一是SRI-ADV数据集，
%包含两个task,
%一个是预测任务，
%一个是生成任务；
%第二是 video conversation benchmark;

%Intra Testing on SRI-ADV：
%如3.4章节所描述，SRI-ADV分为Subjectivity与Objectivity两项任务。重点：要说明数据集有难度，直接做zero-shot效果不好，开源可finetune的方法加入我们的训练之后会有明显的改善，我们自己的方法效果最好；
%Subjectivity：参考3.4章节描述，此项任务包含General User与User Profile两类协议，使用ACC和F1 score两项metric进行评估，如表4所示结果，我们在这两选协议上选择了Gemini-pro-vision，GPT4V，Video-LLaVA，Video-Chat2等著名的多模态大模型方法作为对比方法，我们参照以下步骤进行实验，1. 使用随机函数生成结果，结果分布与测试集分布基本一致，有xx%浮动；2. 在测试集上做zero-shot推理，各个模型的prompt参考补充材料，其中Video-Chat2由于不听从指令，难以得到具备参考意义的结果，其他方法的zero-shot结果均基本高于随机结果，只有Gemini-pro-vision在Emotion指标的预测上出现低于随机结果的情况；3. 使用SRI-ADV提供的训练集，对开源模型Video-LLaVa与Video-Chat2进行finetune，finetune后结果均优于对应的zero-shot结果，Video-Chat2指令听从能力改进明显；4. finetune我们提出的HMLLM方法,对比Video-Chat2，P1上的Engagement,Emotion,EMR Duration三项指标的ACC提升了x%,x%,x%, P2上这三项指标的ACC提升了x%,x%,x%；
%Objectivity:参考3.4章节描述，这项任务的GT是在GPT4Vision标注的基础上进行人工修正的，所以GPT4Vision的zero-shot推理结果相对较高，因Gemini-pro-vision以及GPT4Vision本身不支持视频输入，所以在测试中装备了FSVR，使这两个方法能够具备视频输入的能力。Video-LLaVA在该项任务中fine-tune之后ACC提升了xx%.Score提升了xx%，Video-LLaVA在该项任务中fine-tune之后ACC提升了xx%.Score提升了xx%，装备HMLLM之后ACC提升了xx%.Score提升了xx%；
%video conversation benchmark：目前英文版就可以，还少一部分内容是：我们自己的方法在这个benchmark上也达到了更好的效果（还缺少一个实验结果）；

%Ablation Study：消融实验部分，主要消融lambda;
%Analysis and Visualization：展示几个有代表的QA连续问答对，做一张图，分析这些案例；

\begin{table*}[h]
\centering
\caption{Results of different models on Subjectivity task (Engagement, Emotion, and EMR Duration). Using the Frame Sequence for Video Representation (FSVR) strategy is denoted by a "$\triangle$".}
\vspace{-2mm}
\label{tab:task1-results}
\resizebox{\textwidth}{!}{
\begin{tabular}{lcccccccc}
\toprule
\multirow{2}{*}{ \textbf{Models} } & \multirow{2}{*}{ \textbf{Protocol} } & \multirow{2}{*}{ \textbf{Settings} } & \multicolumn{2}{c}{\textbf{Engagement (2 classes)}} & \multicolumn{2}{c}{\textbf{Emotion (3 classes)}} & \multicolumn{2}{c}{\textbf{EMR Duration (3 classes)}} \\
\cmidrule(lr){4-5} \cmidrule(lr){6-7} \cmidrule(lr){8-9}
& & & \textbf{Acc} & \textbf{F1} & \textbf{Acc} & \textbf{F1} & \textbf{Acc} & \textbf{F1} \\
\midrule
\multirow{2}{*}{ \textbf{Random} } & P1 & --- & 50.44 & 49.93 & 32.30 & 26.26 & 35.01 & 32.10 \\
 & P2 & --- & 50.14 & 50.00 & 33.13 & 33.03 & 33.52 & 33.18 \\
\hline
\multirow{2}{*}{ \textbf{GPT4V$^\triangle$} ~\cite{achiam2023gpt4}} & P1 & Zero-shot & 58.57 & 71.95 & 52.46 & 50.67 & 49.94 & 53.43 \\
 & P2 & Zero-shot & 45.62 & 61.53 & 36.40 & 43.65 & 39.39 & 47.04 \\
\hline
\multirow{2}{*}{ \textbf{Gemini-pro-vision$^\triangle$} ~\cite{team2023gemini}} & P1 & Zero-shot & 59.89 & 73.70 & 17.66 & 20.00 & 46.40 & 47.96 \\
 & P2 & Zero-shot & 46.16 & 63.31 & 30.56 & 43.10 & 36.20 & 43.96 \\
\hline
\multirow{2}{*}{ \textbf{Video-LLaVA} ~\cite{Video-llava}} & P1 & Zero-shot & 60.06 & 74.50 & 61.39 & 71.30 & 45.26 & 57.48 \\
 & P2 & Zero-shot & 46.38 & 61.38 & 31.04 & 42.71 & 31.56 & 49.30 \\
\hline
% \multirow{2}{*}{ \textbf{Video-Chat2} ~\cite{li2023mvbench} } & P1 & Zero-shot & - & - & - & - & - & - \\
%  & P2 & Zero-shot & - & - & - & - & - & - \\
\hline
\multirow{2}{*}{ \textbf{Video-LLaVA} ~\cite{Video-llava}} & P1 & Finetune & 66.29 & 66.85 & 72.33 & 81.94 & 61.05 & 61.80 \\ 
 & P2 & Finetune & 52.58 & 52.69 & 38.62 & 44.72 & 41.28 & 50.84 \\
\hline
\multirow{2}{*}{ \textbf{Video-Chat2} ~\cite{li2023mvbench}} & P1 & Finetune & 75.34 & 76.95 & 71.36 & 75.78 & 57.39 & 60.80 \\
 & P2 & Finetune & 60.06 & 60.02 & 39.66 & 40.24 & 44.06 & 45.51 \\
\hline
\multirow{2}{*}{ \textbf{HMLLM (Ours)} } & P1 & Finetune & \textbf{78.41} & \textbf{79.26} & \textbf{78.41} & \textbf{84.83} & \textbf{62.05} & \textbf{62.43} \\
 & P2 & Finetune & \textbf{64.43} & \textbf{64.65} & \textbf{43.20} & \textbf{48.84} & \textbf{51.96} & \textbf{56.24} \\
\bottomrule
\end{tabular}
}
\end{table*}

\begin{table}
\centering
\caption{Comparative performance of different models on the Objectivity task. Using the FSVR strategy is denoted by a "$\triangle$". The underline of GPT4V denotes the upper bound. We compute the Accuracy (Acc) and VideoChatGPT-Score (Score) \cite{Maaz2023VideoChatGPT} of the proposed method HMLLM and other compared state-of-the-art methods on testing data.}
\vspace{-2mm}
\label{tab:task2-results}
\scriptsize
\resizebox{\columnwidth}{!}{
\begin{tabular}{lccc}
\toprule
\textbf{Models} & \textbf{Settings} & \textbf{Acc} & \textbf{Score \cite{Maaz2023VideoChatGPT}} \\
\midrule
GPT4V$^\triangle$ & Zero-shot & \underline{84.80} & \underline{3.99} \\
\hline
Gemini-pro-vision$^\triangle$& Zero-shot & 27.15 & 2.35 \\
Video-LLaVA ~\cite{Video-llava} & Zero-shot & 15.20 & 2.06 \\
Video-Chat2 ~\cite{li2023mvbench} & Zero-shot & 21.80 & 2.11 \\
\hline
Video-LLaVA ~\cite{Video-llava} & Finetune & 44.76 & 3.03 \\
Video-Chat2 ~\cite{li2023mvbench} & Finetune & 49.27 & 3.12 \\
\textbf{HMLLM (Ours)} & Finetune & \textbf{50.52} & \textbf{3.13} \\
\bottomrule
\end{tabular}
}
\vspace{-2mm}
\end{table}

\begin{table}[h]
\centering
\caption{Results of video conversation benchmark \cite{Maaz2023VideoChatGPT}. CI: Correctness of Information, DO: Detail Orientation, CU: Contextual Understanding, TU: Temporal Understanding, C: Consistency.}
\label{tab:VCB}
\vspace{-2mm}
\resizebox{\columnwidth}{!}{
\begin{tabular}{lcccccc}
\toprule
\textbf{Models}      & \textbf{CI} & \textbf{DO} & \textbf{CU} & \textbf{TU} & \textbf{C} & \textbf{Avg.} \\ 
\midrule
Video LLaMA~\cite{videollama}   & 1.96  & 2.18  & 2.16  & 1.82  & 1.79  & 1.98 \\
Video Chat~\cite{videochat}     & 2.23  & 2.50  & 2.53  & 1.94  & 2.24  & 2.29 \\
LLaMA Adapter~\cite{zhang2023llama}       & 2.03  & 2.32  & 2.30  & 1.98  & 2.15  & 2.16 \\
Video-ChatGPT~\cite{Maaz2023VideoChatGPT}    & 2.40  & 2.52  & 2.62  & 1.98  & 2.37  & 2.38 \\ 
Video-Chat2~\cite{li2023mvbench}        & 3.02  & \textbf{2.88}  & 3.51  & \textbf{2.66}  & 2.81  & 2.98 \\
%Video-Chat2*        & 2.96  & 2.81  & \textbf{3.52}  & 2.48  & 2.85  & 2.92 \\
\textbf{HMLLM (Ours)}& \textbf{3.12} & 2.86 & \textbf{3.52} & 2.61 & \textbf{2.91} & \textbf{2.99} \\
\bottomrule
\end{tabular}
}
\label{VCB-tab}
\vspace{-2mm}
\end{table}

\begin{table}
\centering
\caption{Results of $\lambda$ on Procotol2 of the Subjectivity Task. }
\label{tab:abl-loss-results}
\resizebox{\columnwidth}{!}{
\begin{tabular}{lcccccc}
\toprule
% \multirow{2}{*}{ \textbf{$\lambda_{SALM}$} } &
\multirow{2}{*}{ 
 \textbf{$\lambda$} }  & \multicolumn{2}{c}{\textbf{Engagement}} & \multicolumn{2}{c}{\textbf{Emotion}} & \multicolumn{2}{c}{\textbf{EMR}} \\
\cmidrule(lr){2-3} \cmidrule(lr){4-5} \cmidrule(lr){6-7}
& \textbf{ACC} & \textbf{F1} & \textbf{ACC} & \textbf{F1} & \textbf{ACC} & \textbf{F1} \\
\midrule
% \multirow{2}{*}{ \textbf{1.0} } & \multirow{2}{*}{ \textbf{0.1} }
% \textbf{0.0} & 56.74 & 56.70 & 38.95 & 44.94 & 44.06 & 45.51 \\
\textbf{0.0} & 60.06 & 60.02 & 39.66 & 40.24 & 44.06 & 45.51 \\
\textbf{0.05} & 62.69 & 63.66 & 42.62 & 42.46 & 52.22 & 54.78 \\
\textbf{0.1} & \textbf{64.43} & \textbf{64.65} & \textbf{43.20} & \textbf{48.84} & \textbf{51.96} & \textbf{56.24} \\
\textbf{0.2} & 62.38 & 63.28 & 38.06 & 39.79 & 50.57 & 50.56 \\
\textbf{0.5} & 60.86 & 60.83 & 40.17 & 42.34 & 50.89 & 52.47 \\
\bottomrule
\end{tabular}
}
\vspace{-2mm}
\end{table}

% \subsection{Analysis and Visualization}

\begin{figure*}[h!]
    \centering
    \includegraphics[width=1.0\linewidth]{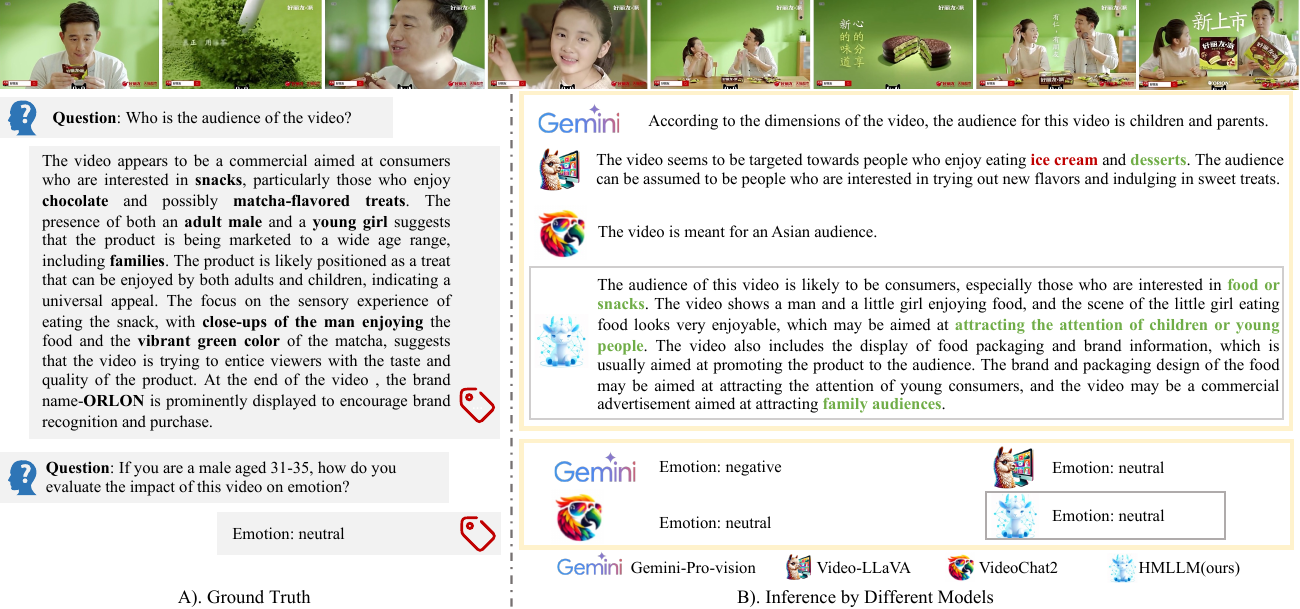}
    \caption{
    Qualitative analysis of SRI-ADV. Green signifies accurate descriptions, while red denotes incorrect responses.
    }
    \vspace{-1mm}
    \label{fig:qual}
\end{figure*}

\noindent\textbf{Metrics.} 
In our study, the Subjectivity Task of SRI-ADV is structured in a multiple-choice question (\textbf{MC}) format. To evaluate its performance, we employ Accuracy (\textbf{Acc}) and F1 score as our metrics. For the zero-shot evaluation of the Subjectivity Task, we have devised a unique prompt, with detailed information provided in the appendix. The subsequent task, named Objectivity, involves open-ended (\textbf{OE}) text generation. For its evaluation, we introduce evaluation measures \cite{Maaz2023VideoChatGPT} based on assessments using GPT-3.5 Turbo.

% \textbf{Implementation Details.}
\noindent\textbf{Implementation Details.}
We employ UMT-L \cite{li2023unmasked} as the
visual encoder and Vicuna-7B-v0~\cite{chiang2023vicuna} as the base model for the SRI-Aware Language Model (SALM). Following the methodology of BLIP2 \cite{blip2}, QFormer is configured with 64 queries. Throughout both the inference and training phases, we adhere to the \textbf{FSVR} strategy detailed
in Section \ref{sec:FSVR}, which involves representing each video with 8 key frames.
% ensuring a representation of 8 keyframes per video. 
%Further details can be found in the appendix.
Further details of hypergraph construction can be found in the appendix.

\vspace{-2mm}
\subsection{Intra Testing} %Validation of the SRI-ADV Dataset}

% \noindent\textbf{Intra Testing on SRI-ADV}
\subsubsection{Intra Testing on SRI-ADV} 
As described in Section \ref{sec:protocol}, we have designed two tasks on the collected SRI-ADV dataset, namely Subjectivity and Objectivity.

\noindent\textbf{Subjectivity task}.
In this task, two protocols are encompassed, \textit{i.e.}, P1, and P2. As shown in Table~\ref{tab:task1-results}, we present our proposed method HMLLM alongside those from renowned MLLMs such as Gemini-pro-vision, GPT4V, Video-LLaVA, and Video-Chat2.
% Based on the results, we can observe from top to bottom:
The results, observed from top to bottom, indicate:
%1) In our experimental framework, the initial step involved generating baseline results through a random function, which serves as the bottom bound.The outcome distribution generated by the random function was designed to closely mirror the distribution of the test set, ensuring a relevant and fair baseline for evaluating model performance.
    
1) For zero-shot inference, GPT4V and Video-LLaVA significantly outperformed the random across-the-board. Gemini-pro-vision underperformed the Random baseline in predicting the accuracy of Engagement and Emotion. Video-Chat2's failure to follow instructions made it difficult to obtain meaningful results. The settings and prompts of zero-shot inference can be found in the appendix.

2) Upon fine-tuning the models with the SRI-ADV dataset, we observed notable improvements in performance for both Video-LLaVA and Video-Chat2 across both P1 and P2, compared to their initial zero-shot configurations. 
% compared to their zero-shot configurations 
% Alternatively, the most significant advancements were demonstrated by our proposed HMLLM method, which consistently outperformed other state-of-the-art methods across all metrics and protocols.
Moreover, our proposed HMLLM demonstrated the most substantial advancements, consistently outperforming other state-of-the-art methods across all evaluated metrics and protocols.

In detail, within Protocol 1, HMLLM surpassed the leading benchmarks in the categories of Engagement (2 classes), Emotion (3 classes), and EMR Duration (3 classes). The improvements were remarkable, showing enhancements in (accuracy, F1) scores by (3.07, 2.31), (6.08, 2.89), and (1.00, 0.63), respectively.
% \yk{(}accuracy and F1\yk{)} 
% 3.07, 2.31; 6.08, 2.89; and 1.00, 0.63, 
These results underscored the efficacy of our method in accurately capturing and analyzing both engagement and emotional dynamics, as well as predicting EMR duration with high precision.
For Protocol 2, the superiority of HMLLM is equally evident. Again, it outshoned the best-existing benchmarks in Engagement (2 classes), Emotion (3 classes), and EMR Duration (3 classes), with enhancements in (accuracy, F1) scores by (4.37, 1.34), (3.54, 4.12), and (7.90, 5.40), respectively. 
These findings highlight the robustness and adaptability of our model across different protocols, further establishing its potential for widespread applicability in real-world scenarios.
%These results collectively affirm the superior performance of the HMLLM method, demonstrating its exceptional capability in handling complex multi-modal data analysis tasks. The consistent outperformance across diverse metrics and protocols not only validates the effectiveness of our approach but also sets a new standard for future research in this domain.

\noindent\textbf{Objectivity Task.}
In the exploration of the objectivity task, as detailed in Section \ref{sec:AVQA}, we meticulously refined the ground truth (GT) by manually correcting annotations initially provided by GPT4V. This meticulous process contributed to the notably high zero-shot inference capabilities observed for GPT4V.
Given that Gemini-pro-vision and GPT4V inherently lack support for video inputs, we integrated Frame Sequence Video Representation (FSVR) technology to bridge this gap. This adaptation endowed both models with the ability to process video inputs, thus expanding their applicability across a wider range of tasks.
% The outcomes of this task, as presented in Table~\ref{tab:task2-results}, offer insightful observations. 
As shown in Table~\ref{tab:task2-results}, GTP4V became the upper bound in a zero-shot setting because we semi-automatically utilized it for labeling, as described in Section \ref{sec:AVQA}.
When the narrative shifts upon the fine-tuning of our models with the SRI-ADV dataset. Both Video-LLaVA and Video-Chat2 showcased substantial enhancements in their performance metrics, surpassing their initial zero-shot configurations. This improvement highlights the transformative impact of targeted training on model efficacy.
% Most notably, 
Notably, our proposed HMLLM method emerged as a formidable contender, eclipsing other models in performance across the board. 
Specifically, HMLLM outperformed the best baseline, Video-Chat2, in terms of accuracy and the Score \cite{Maaz2023VideoChatGPT} by 1.25 and 0.01, respectively. 
%Despite these impressive gains, it is important to acknowledge that our method still trails behind the upper benchmark set by GPT4V by 34.28 and 0.84, respectively.
%This nuanced analysis underscores the competitive landscape of current AI models in handling objectivity tasks. While our HMLLM method sets a new benchmark in model performance, the comparison with GPT4V illuminates the ongoing challenges and the potential for further advancements in the field.

%The results not only validate the effectiveness of fine-tuning with the SRI-ADV dataset but also emphasize the relentless pursuit of excellence in AI research, striving to bridge the gap with the upper bounds of current capabilities.

The results not only validate the effectiveness of fine-tuning with the SRI-ADV dataset but also emphasize that our HMLLM method sets a new benchmark in model performance.

% \noindent\textbf{Intra Testing on Video Conversation Benchmark.} 

\subsubsection{Intra Testing on Video Conversation Benchmark}

%To substantiate the validity of the proposed Electroencephalographic Indices Response to Advertising Videos dataset (SRI-ADV), we conducted evaluations on video benchmarks. Video-ChatGPT\cite{Maaz2023VideoChatGPT} generate task-specific question-answers by querying the GPT-3.5-Turbo model using the human-generated detailed video descriptions. We employed a variety of publicly available models for fine-tuning training based on the SRI-ADV dataset to observe enhancements in performance metrics. As shown in the last row of Table \ref{tab:VCB}, we fine-tuned the Stage 2 and Stage 3 phases of VideoChat2\cite{li2023mvbench} using the SRI-ADV dataset. Experimental results demonstrate that our dataset effectively enhances both Contextual Understanding and Consistency. Given our dataset does not overemphasize temporal details, a slight decrease in Temporal Understanding was observed. 
To further validate the performance of HMLLM, we conducted experiments on other video-based generative performance benchmarks. Following the setup of Video-ChatGPT\cite{Maaz2023VideoChatGPT}, 
we present the performance of our proposed HMLLM, detailed in the last row of Table \ref{tab:VCB}. Experimental results demonstrate that the HMLLM effectively enhances both Contextual Understanding and Consistency. 
Given the HMLLM did not overemphasize temporal details, a slight decrease in Temporal Understanding was observed.

\subsection{Ablation Study}
\noindent\textbf{Effect of $\lambda$.} In the course of training HMLLM, a series of ablation studies were carried out on the $\lambda$ in Equation \ref{eq:loss}, the results of which are detailed in Table \ref{tab:abl-loss-results}. The integration of the SAL-HL Module significantly bolstered the model's proficiency in capturing subjective metrics, culminating in optimal performance at a $\lambda$ value of $0.1$. Beyond this threshold, any further increase in $\lambda$ resulted in a slight decrease in performance, 
likely due to an overemphasis on SAL-HL features at the expense of the SALM's inferential capabilities. Despite this, HMLLM consistently surpasses the baseline model ($\lambda=0.0$) in terms of inference strength, demonstrating the beneficial impact of the hypergraph integration on the model's overall performance.

\subsection{Analysis and Visualization}

We further present a qualitative comparison in Figure \ref{fig:qual}. HMLLM demonstrates an enhanced ability to generate longer and more comprehensive responses for Objectivity Tasks. This improvement can be attributed to the longer average context length of our dataset, which facilitates a deeper understanding of video content by enabling detailed analysis of advertising plots and visual elements. 
%, while also minimizing the introduction of erroneous information. 
% Simultaneously, in subjectivity Tasks, HMLLM exhibits a proficient capability in analyzing videos and inferring subjective indicators of audience engagement and perception. 
More detailed qualitative analyses are available in the appendix.

%% file: sec/06-conclusion.tex
\section{Conclusion}
In this paper, we released a large-scale SRI-ADV dataset with two challenging tasks. We hope it will push cutting-edge research in video understanding. 
%we proposed a novel HMLLM approach can bridge semantic gaps across modalities. 
Besides, we proposed a novel HMLLM approach that enhances the language model by constructing a hypergraph feature space across modalities, thereby providing semantically richer associative features.
Finally, we conducted a comprehensive set of experiments on both SRI-ADV and other video-based generative datasets, verifying the significance of the proposed dataset and method.
% performance benchmark

%% file: sec/07-acknowledge.tex
\section{Acknowledgments}
This work was supported by the Brain-like General Vision Model and Applications project (Grant No. 2022ZD0160403), China Postdoctoral Science Foundation (2023M740079, GZC20230058).
% performance benchmark

%% file: appendix.tex
\newpage
\appendix
% \appendixautorefname{:Hypergraph Multi-modal Large Language Model: Exploiting EEG and Eye-tracking Modalities to Evaluate Heterogeneous Responses for Video Understanding}
% % ==================
\section{Analysis of EEG Raw Signal}

\begin{figure}[h]
    \centering
    \includegraphics[width=0.9\linewidth]{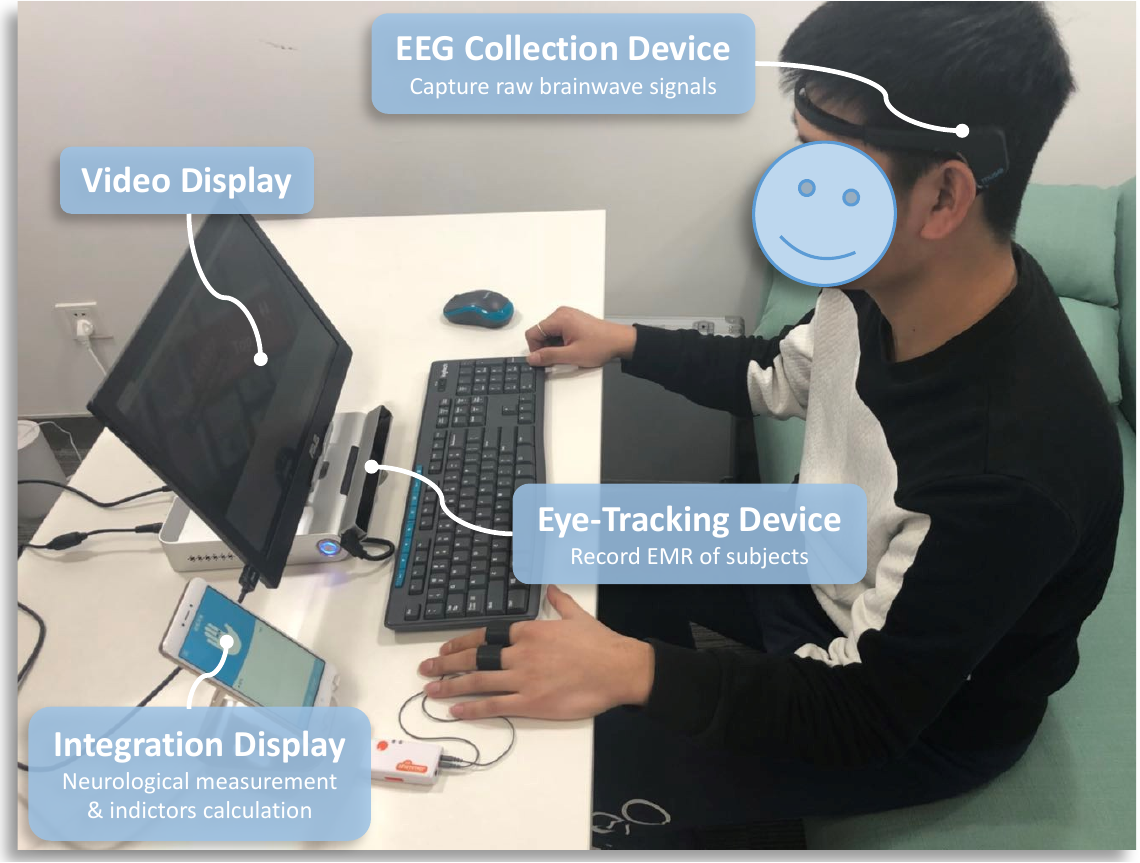}
    \caption{
    Equipment for Collecting Subjective Responses of SRI-ADV dataset. During data acquisition, participants wear an EEG Device, facing a Video Display, with an Eye-Tracking Device below to monitor gaze. Video durations and subjective responses are recorded on an Integration Display for analysis.
    }
    \vspace{-1mm} 
    \label{fig:device}
\end{figure}

Electroencephalography (EEG) stands as a pivotal method for recording the electrical activity of the brain. This is achieved through the placement of electrodes across the scalp, shown in Figure~\ref{fig:device}, to detect electrical signals from neurons. These signals are instrumental in delineating the brain's activity patterns across various cognitive states, providing deep insights into how the brain orchestrates complex psychological emotion\yk{s} and cognitive processes.

% \subsection{Significance of Waves in EEG Signals}
EEG signals are characterized by multiple frequency bands, among which Alpha and Beta waves are paramount, each corresponding to distinct functional states of the brain~\cite{kaur2015eeg, klimesch1999eeg, subasi2021eeg}. We categorize different EEG bands according to the following frequency definitions:
\begin{itemize}
    \item Alpha Waves (8–13 Hz) are emblematic of the brain's state of relaxation and idleness. They are further categorized into three sub-bands based on their frequency range. 1) Alpha1 (8–8.9 Hz): This band is predominant when an individual is in a relaxed state with closed eyes, marking the onset of relaxation. 2) Alpha2 (9–10.9 Hz): These waves are more pronounced when the individual is relaxed yet maintains a level of alertness. 3) Alpha3 (11–12.9 Hz): This band appears as the brain relaxes further while remaining somewhat awake.
    \item Beta waves (13–30 Hz) are integral to the brain's alertness, focused attention, and cognitive processing. 1) Beta1 (13–18 Hz): Associated with mild cognitive activities and focused attention, like reading or simple thought processes. 2) Beta2 (18–22 Hz): These waves intensify during complex cognitive tasks such as problem-solving and decision-making. 3) Beta3 (22–30 Hz): This range signifies highly focused attention and rapid cognitive processing, indicative of active information processing.
% The gradation within Alpha waves from Alpha1 to Alpha3 illustrates the brain's transition from alertness to relaxation, elucidating their significance in cognitive tasks and adaptive processes.
\end{itemize}

% 1. Alpha Waves (8–13 Hz) are emblematic of the brain's state of relaxation and idleness. They are further categorized into three sub-bands based on their frequency range. 1) Alpha1 (8–8.9 Hz): This band is predominant when an individual is in a relaxed state with closed eyes, marking the onset of relaxation. 2) Alpha2 (9–10.9 Hz): These waves are more pronounced when the individual is relaxed yet maintains a level of alertness. 3) Alpha3 (11–12.9 Hz): This band appears as the brain relaxes further while remaining somewhat awake.
% The gradation within Alpha waves from Alpha1 to Alpha3 illustrates the brain's transition from alertness to relaxation, elucidating their significance in cognitive tasks and adaptive processes.

% 2. Beta waves (13–30 Hz) are integral to the brain's alertness, focused attention, and cognitive processing. 1) Beta1 (13–18 Hz): Associated with mild cognitive activities and focused attention, like reading or simple thought processes. 2) Beta2 (18–22 Hz): These waves intensify during complex cognitive tasks such as problem-solving and decision-making. 3) Beta3 (22–30 Hz): This range signifies highly focused attention and rapid cognitive processing, indicative of active information processing.

% In this research, we investigate the analysis of EEG signals and the gathering of eye movement metrics to reveal 
% % the 
% authentic subjective feedback from various demographic groups in response to advertisements. 
In this research, we analyze EEG signals and eye movement metrics to capture authentic subjective feedback from diverse demographic groups in response to advertisements.
The SRI-ADV dataset, derived from raw signals and specific frequency bands, aids in improving the understanding of videos, especially those related to advertising.

% \subsection{Application in Neuromarketing}
% Analyzing EEG signals, particularly in contexts such as advertisement viewing, can reveal insights into cognitive engagement and emotional responses. For example, an uptick in Beta waves might indicate heightened focused attention and cognitive activity in response to engaging content. Conversely, variations in Alpha waves can shed light on the emotional states of relaxation or tension, offering a gauge for the emotional impact of advertisements.

% Moreover, integrating eye movement data with EEG signals can further elucidate the extent of attention paid to specific advertisement elements. This amalgamation of visual attention and brain activity data empowers researchers and advertisers to refine advertisement content and design, optimizing their impact and effectiveness.

% ==================
\section{Data distribution of Subjectivity Task in SRI-ADV}
\begin{table}[t]
\centering
\caption{Categories \& Distribution of Subjectivity Task}
\vspace{-2mm}
\label{tab:SRI-ADV-dis}
\resizebox{\columnwidth}{!}{

\begin{tabular}{c|c|c|c}
\toprule
\textbf{Task} & \textbf{Engagement} & \textbf{Emotion} & \textbf{EMR} \\ 
\midrule
\multirow{3}{*}{\textbf{Cls-1}} & non-cognitive & negative & not attended \\ 
 & $[0, 1)$ & $(-\infty, -6)$ & $[0,0.45)$ \\
 & prop. 55.0\% & prop. 29.7\% & prop. 29.3\% \\ 
\midrule
\multirow{3}{*}{\textbf{Cls-2}} & cognitive & neutral & partially attended \\ 
 & $[1, +\infty)$ & $(-6,6)$ & $[0.45,0.6)$  \\ 
 & prop. 45.0\% & prop. 39.0\% & prop. 41.6\% \\
\midrule
\multirow{3}{*}{\textbf{Cls-3}} & \multirow{3}{*}{--} & positive & fully attended \\ 
 &  & $(6, +\infty)$ & $[0.6, +\infty)$ \\
 &  & prop. 31.2\% & prop. 29.0\% \\ 
\midrule
% \textbf{Distri.} & \includegraphics[width=0.11\textwidth]{figures/dis-label-fig0.png} & \includegraphics[width=0.11\textwidth]{figures/dis-label-fig1.png} & \includegraphics[width=0.11\textwidth]{figures/dis-label-fig2.png} \\
\textbf{Distri.} & \raisebox{-0.45\height}{\includegraphics[width=0.11\textwidth]{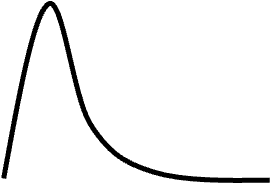}} & \raisebox{-0.45\height}{\includegraphics[width=0.11\textwidth]{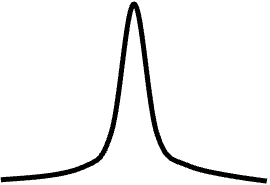}} & \raisebox{-0.45\height}{\includegraphics[width=0.11\textwidth]{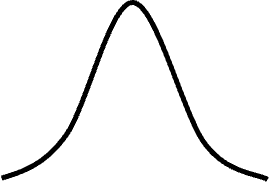}} \\
\bottomrule
\end{tabular}
}
\vspace{-3mm}
\end{table}

We initiate our analysis by performing a comprehensive histogram distribution assessment of Engagement, Emotion, and EMR indicators, anchored to Audience Profiles benchmarks. As delineated in Table \ref{tab:SRI-ADV-dis}, the concluding row graphically encapsulates the distribution of each evaluated SRI. The table's initial three rows delineate each category's designation, value range, and the proportion of data attributed to the 
% respective 
corresponding category. For instance, the Engagement indicator is bifurcated into two categories: the first, termed "non-cognitive," spans a value range from 0 to 1 (non-inclusive), with 55.0\% of observations classified under this category.

Leveraging the statistical insights derived, we proceed to discretize the extant SRI values into distinct categories, thereby facilitating the structuring of the Subjectivity Task associated with SRI-ADV. 

% ==================
\section{Method Algorithm and Training Hyperparameters}
%We now briefly introduce the detail method procedure of this research.
We introduce the Hypergraph Multimodal Large Language Model (HMLLM), a novel approach designed to integrate and process multi-modal data, taking into full account both subjective and objective elements to comprehend advertising videos. Our training procedure is detailed in Algorithm~\ref{alg:alg_pseudo}.
% We present the Hypergraph Multi-modal Large Language Model (HMLLM), a novel approach designed to integrate and process multi-modal data, specifically combining subjective and objective content for advertisement video understanding. 
The method is grounded in the utilization of hypergraphs and %a specialized architecture 
large language model to effectively handle complex relationships within and across modalities. 

We divide the training process into the following stages and 
% adapted 
adapt appropriate training hyperparameters, as shown in Table~\ref{tab:TH}. 

\begin{algorithm}[ht]
\caption{HMLLM: Hypergraph Multi-modal Large Language Model}
\label{alg:alg_pseudo}
\SetKwInOut{To}{to}
\SetKwInOut{Input}{Input}
\SetKwInOut{Output}{Output}
\SetKwInOut{Initialization}{Initialization}
\SetKwData{Video}{V}
\SetKwData{TextualPrompts}{T}
\SetKwData{InitialVisualFeatures}{Fv}
\SetKwData{ProjectedFrameRepresentations}{Fp}
\SetKwData{HypergraphStructure}{G}
\SetKwFunction{ExtractKeyFrames}{ExtractFrames}
\SetKwFunction{PreprocessFrames}{Preprocess}
\SetKwFunction{VisualEncoder}{VisualEncode}
\SetKwFunction{QFormer}{QForm}
\SetKwFunction{SALMProjector}{SALMProjector}
\SetKwFunction{SALMTrain}{SALMTrain}
\SetKwFunction{HypergraphLearning}{HGLearning}
\SetKwFunction{FineTuneModel}{FineTune}
\SetKwFunction{GenerateText}{Generate}
\SetKwFunction{ConcatenateFeatures}{Concatenate}
\SetKwFunction{PoolFeatures}{Pool}
\SetKwFunction{ConstructHypergraph}{Construct}
\SetKwFunction{CombineLosses}{Combine}
\SetKwFunction{InitializeQuerySet}{InitializeQuery}

\Input{
$\text{Video Key Frames} \ F = \{ f_0, f_1, \cdots, f_N \}$, \\
$\text{Textual Prompts} \ T$, \\
$\text{Ground Truth} \ \mathbf{Y}_{gt}$, \\
$\text{Warm Up Epoch} \ E_0$, \\
$\text{Fine-tune Epoch} \ E_1$, \\
}

% Initialization
\Initialization{
$\mathbf{X}_0 \gets \text{Pre\_process}(F_v)$\\
$\mathbf{F}_v \gets \text{Visual\_Encoder}(\mathbf{X}_0)$\\
$\text{SALM} \gets \text{Initialize\_SALM}(\mathbf{F}_v, T)$\\
$\text{HL-Gate} \gets \text{OFF}$\\
$\mathbf{Q} \gets \text{Initialize\_Query}()$
}

\tcp{Stage I: SALM Warm Up}
\For{$i \gets 1 \ \text{to} \ E_0$}{
    $\mathbf{K}, \mathbf{V} \gets \text{QFormer}(\mathbf{F}_v)$\\
    $\mathbf{F}_p \gets \text{SALM\_Projector}(\mathbf{F}_v, \mathbf{K}, \mathbf{V})$\\
    $\text{SALM} \gets \text{SALM\_Train}(\text{SALM}, \mathbf{F}_p, T)$\\
    $\widetilde{\mathcal{Y}}_{qa} \gets \text{SALM}(\mathbf{F}_p, T) $ \\
    $\mathcal{L}_{ITG} \gets \text{ITG\_Loss}(\widetilde{\mathcal{Y}}_{qa}, \mathbf{Y}_{gt})$\\
    $\text{SALM\_Optimizer}(\mathcal{L}_{ITG})$\\
}

\tcp{Stage II: SAL-HL Fine-tuning}
$\text{HL-Gate} \gets SRI\_Contained(T)$ \ 
\tcp{Set HL-Gate ON.}
\For{$i \gets 1 \ \text{to} \ E_1$}{
    $\mathbf{F}_{pv} \gets \text{Feature\_Mixer}(\mathbf{F}_p, \mathbf{F}_v)$\\
    $\mathbf{F}_{frame\_level} \gets \text{Pool}(\mathbf{F}_{pv})$\\
    $\mathcal{G} \gets \text{Construct\_Hypergraph}(\mathbf{F}_{frame\_level}, \mathfrak{K})$\\
    $\widetilde{\mathcal{Y}}_{sri} \gets \text{HGNN}(\mathcal{G},  \mathbf{F}_{frame\_level})$\\
    $\mathcal{L}_{CE} \gets \text{Cross\_Entropy}(\widetilde{\mathcal{Y}}_{sri}, {\mathcal{Y}}_{sri})$\\
    \tcp{ $\mathcal{L}_{ITG}$ is obtained same with the stage I}
    $\mathcal{L} \gets \text{Combined\_Loss}(\mathcal{L}_{ITG}, \mathcal{L}_{CE}, \lambda)$\\
    $\text{Joint\_Optimizer}(\text{SALM}, \text{HGNN}, \mathcal{L})$
}
\end{algorithm}

\subsection{Initialization for Model Parameters}
The process begins with the extraction of key frames from the input video, which are then pre-processed to standardize the input format. These initial visual features are encoded using a visual encoder, producing a set of feature vectors. Simultaneously, textual prompts are prepared for processing. A query set is initialized, marking the starting point for our model's learning process.

\subsection{Stage I: SALM Warm Up}
% GPT生成的把SALM的意思都生成错了
% The Self-Attention Large Model (SALM) Warm Up phase involves the transformation of visual features through a query-former mechanism, which generates key and value pairs for attention mechanisms. These transformed features are then projected using the SALM projector, integrating the textual prompts into the model's understanding. The SALM is trained over a predefined number of warm-up epochs, optimizing an Image-Text Grounding (ITG) loss that aligns the generated outputs with ground truth annotations.
Stage I is dedicated to the training of the SRI-Aware Language Model (SALM), with the goal of enhancing its capabilities in language generation and reasoning inference. In this stage, visual features are converted into key-value pairs using a query-former mechanism, essential for the attention processes. These features are then fed through the SALM projector, which enriches the model's understanding by integrating textual prompts.

The training of SALM spans 10 epochs, as detailed in Table~\ref{tab:TH}, with a particular focus on minimizing the Image-Text Grounding (ITG) loss. This step is crucial for ensuring that the model's outputs are in alignment with the ground truth, thereby optimizing performance.

\subsection{Stage II: SAL-HL Fine-tuning}
% Upon completion of the warm-up phase, the model transitions to fine-tuning with the activation of the Hypergraph Learning (HL) gate. This phase introduces a novel feature mixer that combines the previously obtained projected frame representations with the initial visual features, enriching the model's multi-modal context. The combined features are pooled to a frame-level representation, which serves as the basis for constructing a hypergraph that captures the complex interrelations among the data points. 

Stage II shifts the focus to fine-tuning the SRI-Aware Language Hypergraph Learning (SAL\_HL) component, with the objective of enhancing the model's capability to mimic the subjective perceptual capacities of the brain. After completing the initial warm-up phase, the model enters the fine-tuning stage, marked by the activation of the Hypergraph Learning (HL) gate. This process enriches the model's multi-modal context by combining these features into a pooled frame-level representation. This representation then forms the foundation for constructing a hypergraph that captures the intricate interconnections among data points. 

% A Hypergraph Neural Network (HGNN) is then applied to the hypergraph structure, enabling the model to learn from the intricate connections within the data. The output of the HGNN is fine-tuned against a Cross-Entropy loss, alongside the ITG loss carried over from the warm-up phase. The combined loss function guides the optimization of both the SALM and the HGNN components, ensuring coherent learning across both stages of the model's training.
Following the construction of the hypergraph, a Hypergraph Neural Network (HGNN) is employed to process the hypergraph structure, allowing the model to leverage the complex connections present within the data. 
% The output generated by the HGNN undergoes fine-tuning, utilizing a Cross-Entropy loss in conjunction with the Image-Text Grounding (ITG) loss that was emphasized during the warm-up phase. 
The output generated by the HGNN is fine-tuned using Cross-Entropy loss along with the Image-Text Grounding (ITG) loss emphasized during the warm-up phase.
This amalgamation of loss functions serves as a directive for the optimization process, targeting both the SRI-Aware Language Model (SALM) and the HGNN components. This strategic approach ensures a unified and coherent learning experience throughout the two distinct stages of the model's training, fostering a comprehensive understanding and adaptation to the intricacies of the data.

Following \cite{li2023mvbench}, 
% during Stage 2, 
we incorporate Low-Rank Adaptation (LoRA) \cite{lora} modules into the SALM with a configuration of rank 16, an alpha value of 32, and a dropout rate of 0.1 during Stage II. Within the HGNN, it is imperative to adjust the input based on pooled frame-level representation, setting the Number of Vertices and Hyperedges to \(8 \times 98\) and the Channel of Vertex Representation to 1024. For the hypergraph's internal configuration, we adhere to the commonly used settings as illustrated in the Tabel~\ref{tab:TH}, ensuring a balance between training effectiveness and model size.

\subsection{Summary}
HMLLM stands as a comprehensive framework that leverages the strengths of hypergraph structures and multi-modal data integration. Through its two-stage training process, it achieves a deep understanding of the relationships within and across modalities, paving the way for advanced applications in multi-modal data processing and generation.

\begin{table}[t]
\centering\caption{Training Hyperparameters for different stages.}
\label{tab:TH}
\resizebox{\columnwidth}{!}{

\begin{tabular}{l|cc}
\toprule
\multirow{2}{*}{\textbf{config}} & \textbf{Stage1} & \textbf{Stage2} \\
 & SALM Warm Up & SAL-HL Fine-tune \\
\midrule
input frame & \multicolumn{2}{c}{8} \\
input resolution & \multicolumn{2}{c}{224} \\
max text length & \multicolumn{2}{c}{512} \\
optimizer & \multicolumn{2}{c}{AdamW} \\
optimizer momentum & \multicolumn{2}{c}{$\beta1, \beta2=0.9, 0.999$} \\
weight decay & \multicolumn{2}{c}{0.02} \\
learning rate schedule & \multicolumn{2}{c}{cosine decay} \\
learning rate & 1e-4 & 2e-5 \\
batch size & 128 & 64 \\
warmup epochs & 0.5 & 1 \\
total epochs & 10 & 20 \\
$\lambda$ of $\mathcal{L}$ & 0 & 0.1 \\
augmentation & \multicolumn{2}{c}{flip, MultiScaleCrop [0.5, 1]} \\
\midrule
Vertices number of Hypergraph & - & 8*98 \\ %\multicolumn{2}{c}{8*98} \\
Hyperedges number of Hypergraph & - & 8*98 \\ %\multicolumn{2}{c}{8*98} \\
Channel of Vertex representation & - & 1024 \\ %\multicolumn{2}{c}{1024} \\
K of hypergraph construction & - & 3, 4, 5 \\ %\multicolumn{2}{c}{3, 4, 5} \\
Average out-degree of vertices & - & 12.5 \\ %\multicolumn{2}{c}{12.5} \\
In-degree of hyperedges & - & 3, 4, 5 \\ %\multicolumn{2}{c}{3, 4, 5} \\
\bottomrule
\end{tabular}
}
\end{table}

\section{Computational Complexity and Resource Utilization of HMLLM}
We have recorded the training time on eight A100 GPUs, each with 40GB of memory, and the inference time on a single A100 GPU with the same specifications.
Our proposed HMLLM is generally on par with other models supporting video multi-modalities in terms of parameter count, training time, and inference time. 

\begin{table}[ht]
    \caption{Comparative Analysis of Model Parameters}
    \label{tab:rub-param}
    \centering
    \resizebox{\columnwidth}{!}{
    \begin{tabular}{c|c|c}
        \toprule
        Model & Video-Chat2 & HMLLM (Ours) \\
        \midrule
        Total Parameters (Billions)  & 7.2 & 7.2 \\
        Training Time (H/Epoch) & 13.5 & 14.3 \\
        Inference Time (s/Video) & 6.2 & 6.2 \\
        ACC on Task2 & 49.27  & 50.52 \\
        \bottomrule
    \end{tabular}
    }
\end{table}

% ==================
  \begin{figure*}[h]
    \centering
    \includegraphics[width=1.0\linewidth]{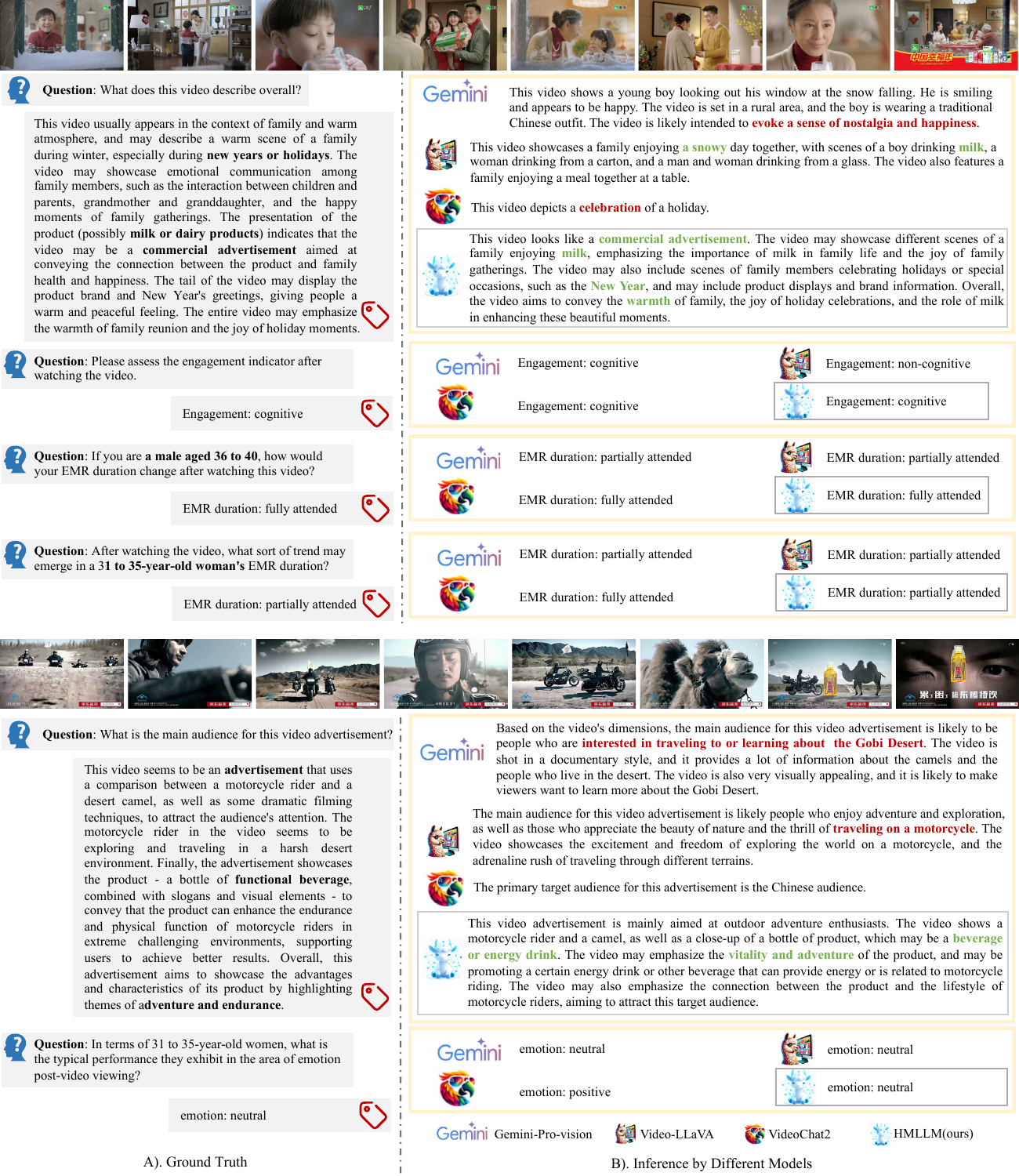}
    \caption{
    More qualitative analysis of SRI-ADV. Green signifies accurate descriptions, while red denotes incorrect or hallucinatory responses.
    }
    \vspace{-1mm} 
    \label{fig:qual}
\end{figure*}

\section{Zero-shot Prompt for Subjectivity Task}
For the subjectivity task, we conducted Zero-shot inference on commercial and open-source models. During this process, we tested various Prompts \footnote{\url{https://clickup.com/blog/ai-prompt-templates/}} \footnote{\url{https://www.zdnet.com/article/how-to-write-better-chatgpt-prompts-in-5-steps/}} to enable reasoning by the language models. The Prompts we selected are as follows.

\newenvironment{specialtext}
{\begingroup\fontfamily{phv}\selectfont\itshape}
{\endgroup}
\begin{specialtext}
As an AI model simulating EEG indicator analysis, your task is to systematically evaluate a user's feelings after viewing video or frames content using the provided EEG indicators: Cognitive Engagement (CE), Emotional Recognition (ER), and Eye Movement Ratio (EMR). 

Effective analysis requires detailed video content information themes, narrative structure, visual and auditory elements and viewer attributes, including age, gender, preferences, and experiences. You will integrate the EEG indicators' definitions and probabilities to deduce the viewer's cognitive engagement, emotional response, and attention level.

This involves assessing how video elements may attract or repel the viewer, grounded in psychological principles and media consumption research.But you don't need to output the reasoning process, only the final result.

1. Cognitive Engagement (CE) Definitions: 

   - A. Non-Cognition:(55\% probability)The viewer shows minimal interest or understanding, resulting in low EEG activity. 
   
   - B. Cognition:(45\% probability)The viewer understands and relates to the video, evidenced by increased EEG activity.

2. Emotional Recognition (ER) Definitions: 

   - A. Negative:(30\% probability)Dislikes certain video elements (e.g., conflict, unappealing objects). 
   
   - B. Neutral:(40\% probability)Feels indifferent towards the video content (e.g., mundane tasks). 
   
   - C. Positive:(30\% probability)Experiences enjoyment or excitement (e.g., appealing scenes). 

3. Eye Movement Ratio (EMR) Definitions:  

   - A. Not Attended:(30\% probability)Viewing ratio $\leq$ 0.45, possibly due to unattractive visuals or cognitive dissonance. 
   
   - B. Partially Attended:(40\% probability)Viewing ratio between 0.45 and 0.6, suggesting some attractive elements. 
   
   - C. Fully Attended:(30\% probability)Viewing ratio > 0.6, indicating high appeal and mood enhancement.

Please ensure your analysis follows this format with no additional output: 

CE: B; 

ER: B; 

EMR: C 

Finally, analyze the specified question without extraneous , focusing on the indicators' specific index based on probabilities.

\{[The Question in Test Set]\}
\end{specialtext}

It is worth noting that to prevent the model from overfitting to specific choices, we randomized the options and tested them three times, taking the average result as the conclusive outcome.

\section{Visualization and Qualitative Analysis}
In the realm of advertising, the use of metaphors, scenic portrayals, and related content is prevalent. Our SRI-ADV dataset is meticulously crafted to support both subjective and objective analyses, thereby offering a comprehensive understanding of video advertising content. It uniquely bridges the gap between these analyses, with objective comprehension bolstering subjective interpretation. This fusion enables the exploration of qualitative aspects such as Engagement, Emotion, and Eye Movement Ratio (EMR) across various demographics.

As shown in Figure~\ref{fig:qual}, Part A showcases the SRI-ADV's ground truth, distinguished by its detailed annotations and extensive response lengths. Meanwhile, Part B delineates a comparative analysis among the outputs generated by the Gemini-Pro-vision, Video-LLaVA, and VideoChat2 models against our HMLLM. 

For instance, an energy drink advertisement as shown in the bottom of Figure~\ref{fig:qual}, HMLLM uniquely captures both the overt (a motorcycle rider and a camel) and the covert (the product's essence of vitality and adventure) elements of the advertisement. This comprehensive analysis extends to the advertisement’s main audience, design principles, visual narratives, and product attributes, showcasing our model's superior capability in extracting and interpreting complex thematic elements.


% \clearpage
% % \newpage
% \bibliographystyle{ACM-Reference-Format}

% \end{document}